\documentclass[10pt,twocolumn,letterpaper]{article}

\usepackage{wacv}
\usepackage{times}
\usepackage{epsfig}
\usepackage{graphicx}
\usepackage{amsmath}
\usepackage{amssymb}
\usepackage{booktabs}
\makeatletter
\@namedef{ver@everyshi.sty}{}
\makeatother
\usepackage{tikz}
\makeatletter

\DeclareMathOperator*{\argmin}{argmin}

\usepackage{subcaption}
\usepackage{gensymb}
\usepackage[linesnumbered,ruled,vlined]{algorithm2e}
\usepackage[noend]{algpseudocode}
\DontPrintSemicolon
\usepackage{array}
\newcolumntype{P}[1]{>{\centering\arraybackslash}p{#1}}

\usepackage[pagebackref=true,breaklinks=true,letterpaper=true,colorlinks,bookmarks=false]{hyperref}
\usepackage[blocks]{authblk}
\makeatletter
\renewcommand\AB@affilsepx{, \protect\Affilfont}
\newcommand{\adhi}{\begin{tikzpicture}
    \draw[line width=0.6mm] (0,0) -- (0,4);
    \end{tikzpicture}}
\makeatother

\def\wacvPaperID{919} 

\wacvalgorithmstrack   

\wacvfinalcopy 

\ifwacvfinal
\usepackage[breaklinks=true,bookmarks=false]{hyperref}
\else
\usepackage[pagebackref=true,breaklinks=true,colorlinks,bookmarks=false]{hyperref}
\fi

\begin{document}

\title{Relation Preserving Triplet Mining for Stabilising the Triplet Loss in Re-identification Systems}


\author[1,2]{Adhiraj Ghosh}
\author[1,3]{Kuruparan Shanmugalingam}
\author[1]{Wen-Yan Lin}
\affil[1]{Singapore Management University}
\affil[2]{University of Tübingen}
\affil[3]{University of New South Wales}


\maketitle
\thispagestyle{empty}

\begin{abstract}
   Object appearances change dramatically with pose variations.
This creates a challenge for embedding schemes that
seek to map instances with the same object ID to locations
that are as close as possible. This issue becomes significantly
heightened in complex computer vision tasks such as
re-identification(reID). In this paper, we suggest that these
dramatic appearance changes are indications that an object
ID is composed of multiple natural groups, and it is
counterproductive to forcefully map instances from different
groups to a common location. This leads us to introduce
Relation Preserving Triplet Mining (RPTM), a feature
matching guided triplet mining scheme, that ensures that
triplets will respect the natural subgroupings within an object
ID. We use this triplet mining mechanism to establish
a pose-aware, well-conditioned triplet loss by implicitly enforcing
view consistency. This allows a single network to be
trained with fixed parameters across datasets, while providing
state-of-the-art results. Code is available at \url{https://github.com/adhirajghosh/RPTM_reid}.
\end{abstract}

\section{Introduction}

Re-identification is the process of identifying images of the same object taken under different conditions. 
One of the main challenges of reID is pose-induced appearance changes~\cite{bai2018group,chu2019vehicle}. 
Not only does object appearance change with pose, different objects often look similar when viewed from the same pose, also known as inverse-variability.
This paper suggests a new interpretation of the inverse-variability problem, one with the potential to significantly improve the effectiveness of reID algorithms. 
Although we focus on re-identification, the underlying principles developed here are not restricted to this task and have the potential to impact a wide range of other computer vision problems~\cite{arandjelovic2016netvlad,lin2021shell,park2019relational,roth2019mic}. 
Current reID frameworks deploy representation and metric learning methodologies in the attempt to learn embeddings that map semantically similar instances to relatively nearby locations; and semantically dissimilar images to relatively distant locations. 
This is typically achieved through a metric loss function such as triplet loss ~\cite{schroff2015facenet}, which encourages a reference (anchor) input to be more similar to  a positive (truthy) input than  to a  negative (falsy) input. 
The number of triplet combinations tend to grow polynomially with the number of instances in a dataset, as detailed by Hermans \textit{et al.} ~\cite{hermans2017defense}; however, most triplet combinations are redundant. 
This has led to the development of triplet mining, whose aim is to identify the most important triplets in a given sample set.  
While triplet mining is ubiquitous in reID algorithms~\cite{bai2018group,he2020multi,hermans2017defense,tang2019pamtri}, it has an innate vulnerability.

\begin{figure}[t]
\begin{center}
   \includegraphics[width=0.9\linewidth]{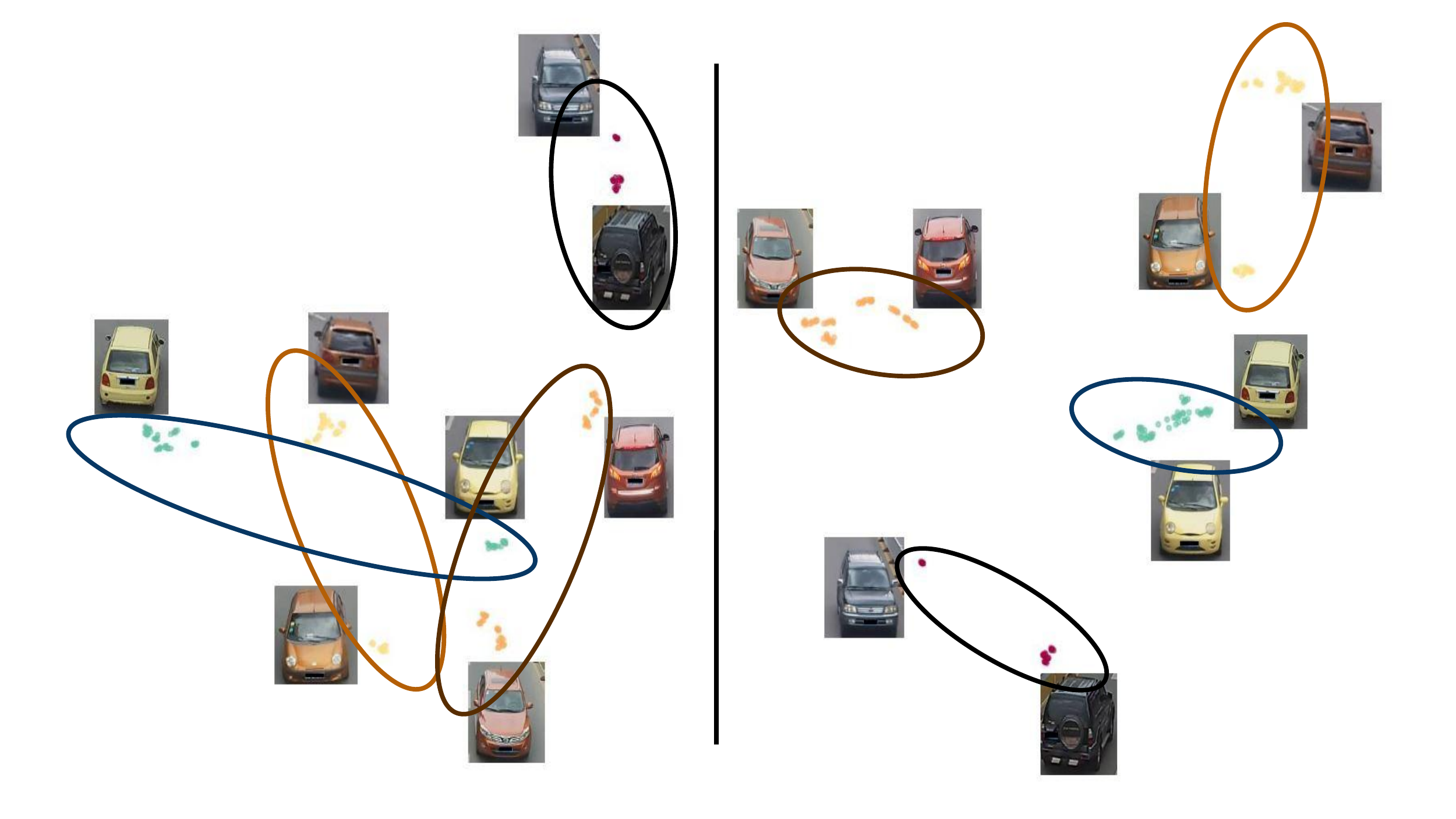}
   \put(-190, 100){\textbf{\underline{DMT}}}
   \put(-90, 100){\textbf{\underline{RPTM}}}
   \put(-109,0){\adhi}
\end{center}
   \caption{Comparing the  features learned by DMT~\cite{he2020multi}, a current state-of-the-art, with our proposed Triplet Mining scheme. Features correspond to the  first four IDs of Veri-776~\cite{liu2016large}. The distance preserving  UMAP  projection shows the  RPTM feature transform is more intuitive.}
\label{fig:umap}
\end{figure}

\begin{figure*}[t]
\begin{center}
     \begin{subfigure}[b]{0.37\linewidth}
         \centering
         \includegraphics[width=\linewidth]{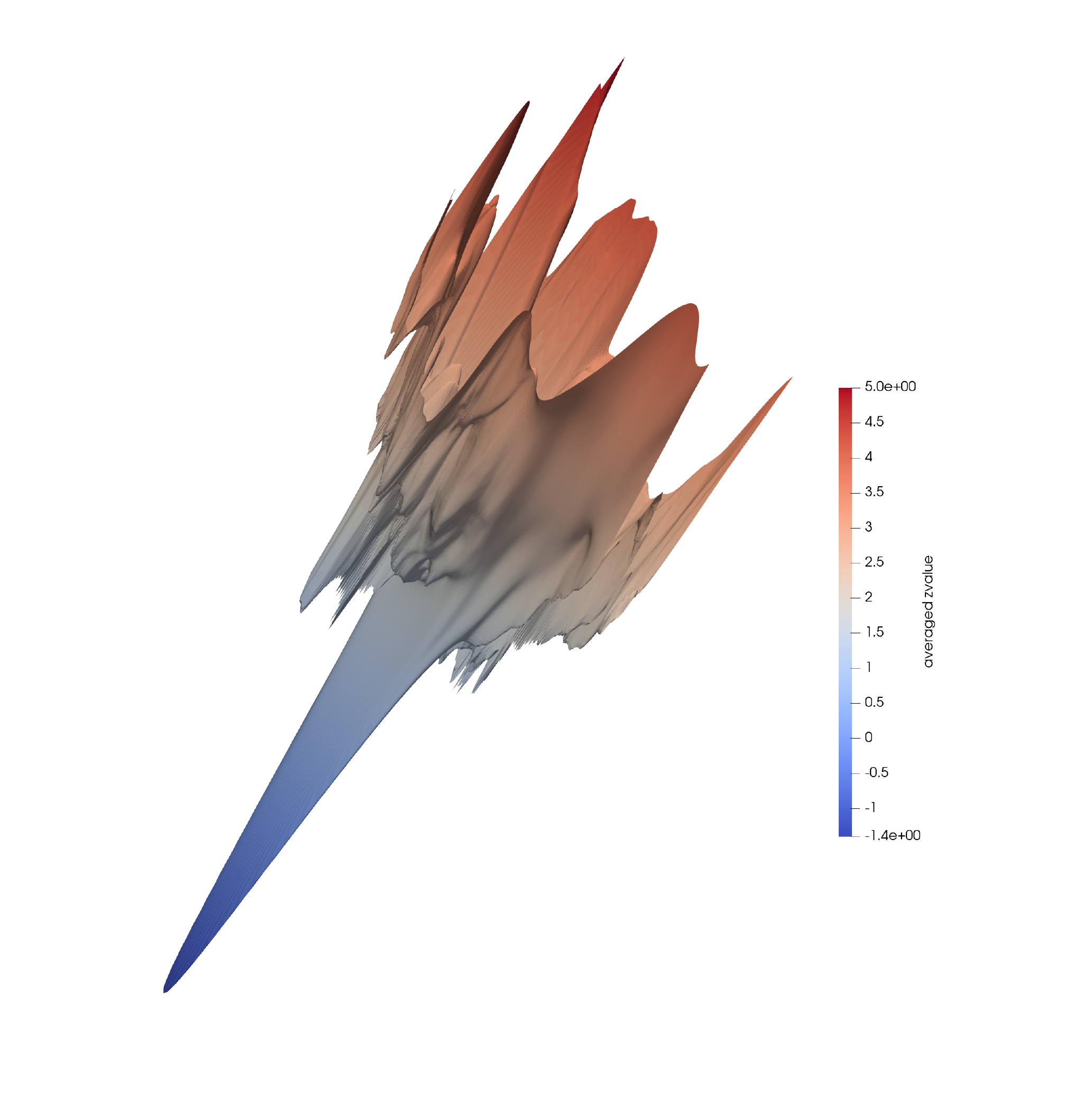}
         \caption{Without RPTM}
         \label{fig:resnet_loss}
     \end{subfigure}
     \begin{subfigure}[b]{0.37\linewidth}
         \centering
         \includegraphics[width=0.8\linewidth]{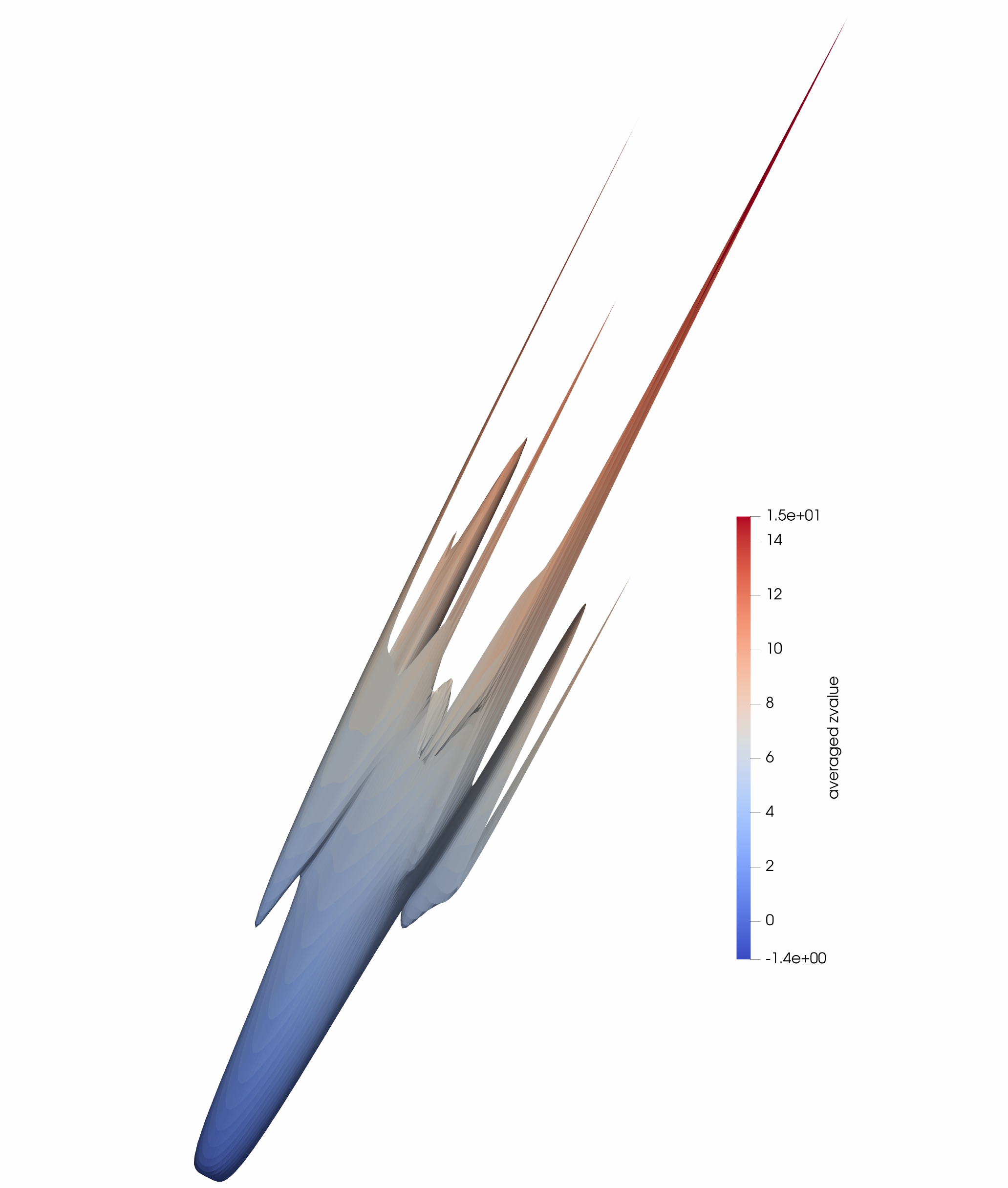}
         \caption{With RPTM}
         \label{fig:rptm_loss}
     \end{subfigure}
     \caption{Loss landscape visualisation of  a ResNet-50 trained with SGD using Triplet Loss on Veri-776 with/without Relation Preserving Triplet Mining. RPTM demonstrates smoother loss surfaces, improved model generalisation and a wider minima, thus allowing better optimisation during training.}
    \label{fig:loss}
\end{center}
\end{figure*}

Consider a hypothetical dataset  containing instances of apple-the-phone and apple-the-fruit, both of which are classified as Apple. The dataset also has instances of phones made by Samsung, classified as Samsung-phone. 
This dataset will have many  difficult triplets, for example, 
apple-the-phone (anchor), apple-the-fruit (positive) and Samsung-phone (negative), which
triplet-mining techniques are encouraged to focus on. However, training with  such triplets is  counter-productive as they attempt to  ensure that instances of apple-the-phone are mapped  closer to instances of apple-the-fruit than  to 
instances of Samsung-phone. 
Such a mapping mechanism violates the natural appearance relation between objects, and it can be seen that current metric learning systems enforce vastly different views of the same object to be coincident in feature space. It is unlikely that models trained on this hypothesis generalise adequately. 

A similar phenomenon occurs in reID, where most datasets~\cite{liu2016deep,liu2016large,ristani2016performance}
group instances by ID. 
However, the appearance of a person or vehicle's front, rear and sides profiles are very different from each other and they appear to belong to physically different entities. 
This creates fallacious anchor-positive pairs, where the instances chosen to be anchor and positive do not share a natural group~\cite{bai2018group}.
This fallacy in the triplet mining scheme can be further realised considering the fact that in \cite{schroff2015facenet}, triplet loss was defined for face detection, where datasets only have the front view of the face, hence all anchor-positive pairs are semantically meaningful. Due to this, triplet mining does not generalise well to reID. 
This problem has been recognised in recent reID works~\cite{chu2019vehicle,khorramshahi2020devil,liu2021spatial,meng2020parsing,tang2019pamtri}, who incorporate pose awareness into the network, and in metric learning~\cite{roth2019mic}, in which latent characteristics shared within and between classes are explicitly learnt.
Although this approach can be effective, it complicates network training and incurs an additional burden of training a new, dataset-specific, pose-aware layer. 

We suggest a simple alternative, where feature matching~\cite{bian2017gms,lowe1999object} is leveraged to discover natural groupings. 
Therefore, we propose \textit{Relation Preserving Triplet Mining} (RPTM), a triplet mining scheme that respects natural appearance groupings. 
We further define our solution as \textit{Implicitly Enforced View Consistency}, which we define as the process of exploiting internal, natural groupings within a class and mapping instances with the same view together as a semantic entity, to overcome intra-class separability.  
These groupings follow natural patterns referenced by semantics~\cite{lin2021shell}, and tend to be pose-related in the context of  reID.  
Here,  RPTM  implicitly enforces pose-aware triplet mining, which prevents  different poses from being  mapped onto one another.
This improves the conditioning of the   triplet-cost, allowing  for the same training  parameters to be employed across a variety of different datasets.  
 The resultant feature embeddings provide better reID results and are more intuitive,  as  shown in Figure \ref{fig:umap}. We observe that past triplet mining processes fail in terms of pose awareness and this may lead to poor ranking results, whereas RPTM not only shows pose awareness, better conditioned triplet mining also ensures accurate ranking results.
 
 Our experiments are structured to demonstrate how a coherent triplet mining scheme can eliminate the largest vulnerability of using triplet loss in reID, without the requirement of key-point labels and pose estimation pipelines. One indicator of the effectiveness of our method is observing the loss optimisation landscape during training. Due to the smooth loss landscape for RPTM, shown in figure \ref{fig:loss}, we demonstrate how RPTM cleans up the triplet mining process with a triplet filtration step and prevents erroneous local minimas. Thus, when trained with RPTM, models with larger parameters can optimise just as fast as smaller networks, which serves our main goal of achieving impressive retrieval results with self-imposed constraints on compute power as well as generalising parameter settings across tasks and datasets. As RPTM is robust to fluctuations of loss landscape, training deeper networks with SGD on a simple cost function is more accessible for object retrieval tasks. 

  In summary, our paper  contributions are:
\begin{enumerate}
    \item We explain how traditional triplet mining methods are  ill-conditioned because it does not take into account natural groupings; 
   \item We propose a feature guided triplet mining scheme that we term Relation Preserving Triplet Mining (RPTM);
  \item We show  RPTM is well-conditioned enough to permit the  use of constant training parameters across  datasets and tasks.  The resultant network is simultaneously capable of  state-of-the-art in  vehicle reID and competitive results for person reID.
\end{enumerate}

\section{Related Works}
\textbf{Re-identification.} 
The demand for urban surveillance applications has led to a surge of interest in person and vehicle re-identification. 
Challenge benchmarks such as VehicleID~\cite{liu2016deep}, Veri-776~\cite{liu2016large}, DukeMTMC~\cite{ristani2016performance} and others have been established; and many new algorithms have been proposed~\cite{he2020multi,khorramshahi2020devil,li2021diverse,liu2021spatial,sun2021tbe,zang2021learning}.
In reID, many algorithms achieve good results by estimating vehicular pose. 
Notably, Tang \textit{et al.}~\cite{tang2019pamtri} created a synthetic data set for pose estimation and Meng \textit{et al.}~\cite{meng2020parsing} used a parser model to split vehicles into four parts for pose-aware feature embedding. 
Recently, Vision Transformers (ViT) for reID ~\cite{he2021transreid,wang2022feature,zhu2022dual} were proposed for attention learning and \cite{Fu_2022_CVPR} addressed person reID with noisy labels. 
We suggest that the root problem encountered by most of these techniques lies in their definition of triplet loss. 
By replacing traditional triplet losses with our RPTM technique, we show that it is possible to achieve state-of-the-art results by minimizing a simple cost function.
This stands out from the trend towards ever more complex reID techniques.

\textbf{Triplet Loss.}
The triplet loss was first introduced in the context of face identification~\cite{schroff2015facenet}. 
Since then, it has undergone many refinements
~\cite{bai2018group,hermans2017defense,xuan2020hard,xuan2020improved}.
Such triplet-based formulations implicitly assume that the given IDs correspond to meaningful groups. 
We suggest that this assumption is often wrong and that triplets should be defined with respect to naturally occurring groups rather than the given labels. 
This perspective on triplet loss differs significantly from that used in most papers. 
To our knowledge, the research most similar to ours is  Bai~\textit{et al.}~\cite{bai2018group} who acknowledge the importance of naturally occurring groups within an ID. 
However, Bai~\textit{et al.} attempts to use the groups to force tighter mappings of an ID,  fighting rather than harnessing the natural relationships. 
Another problem for clustering based works like  Bai~\textit{et al.}~\cite{bai2018group}'s, is that  variations often  have no  naturally occurring cluster boundaries. 
This is not a problem for RPTM which defines relations  in  a pairwise manner, rather than on the basis of shared clusters.

\textbf{Feature Matching.}
RPTM uses feature matching to help establish triplets. 
Feature matching is a well-established field in computer vision, whose goal is to match key points between image pairs. 
Classic feature matching works include SIFT~\cite{lowe1999object},  SURF~\cite{bay2006surf}, ORB~\cite{rublee2011orb}, \textit{etc}.
Recent developments include \cite{bellavia2022sift} for exploiting matching context information, and \cite{ma2022feature} for mismatch removal between two features sets. 
In this paper, we employ Grid-Based Motion Statistics (GMS)~\cite{bian2017gms}  as our feature matcher of choice. 
This is a newer algorithm which incorporates match coherence~\cite{lin2014bilateral} to facilitate  key-point matching.  
GMS outperforms most classic techniques while also being much faster.    

\section{Why Triplet Loss?}

\subsection{Neural Networks as Embedding Functions}

Much of computer vision can be interpreted as an attempt to map 
image instances to a semantically meaningful embedding. Thus,
if $\mathbf{x}_k$ represents an image instance and $\mathbf{y}_k$ its associated feature,
the transformation from $\mathbf{x}_k$ to $\mathbf{y}_k$   can be denoted by $\mathbf{y}_k$ = $f(\mathbf{x}_k)$, where $f: \mathbb{R}^{3 \times w \times h} \to \mathbb{R}^{d} $;   $w \times h$  denotes  image  dimension; 
and  $d$ represents the embedding space's dimensions. In this scheme, the embedding function $f(.)$ is learnt by minimising the cross-entropy loss 
\begin{equation}
\label{eq:ent_final}
E_{ent} = \sum_{k=1}^m \mathcal{L}_{ent}(\mathbf{x}_k),
\end{equation}
where $m$ denotes the total number of training images. 

Minimising the cost in Eq. \ref{eq:ent_final}  provides an embedding that maximises classification accuracy.
 However, this does not ensure that the embedding is semantically meaningful. The retrieval problem
 requires an embedding in which semantically similar instances are 
mapped close to each other, leading to the development of triplet loss~\cite{schroff2015facenet}.

 \begin{figure*}[h]
     \centering
     \begin{subfigure}[b]{0.25\textwidth}
         \centering
         \includegraphics[width=\textwidth]{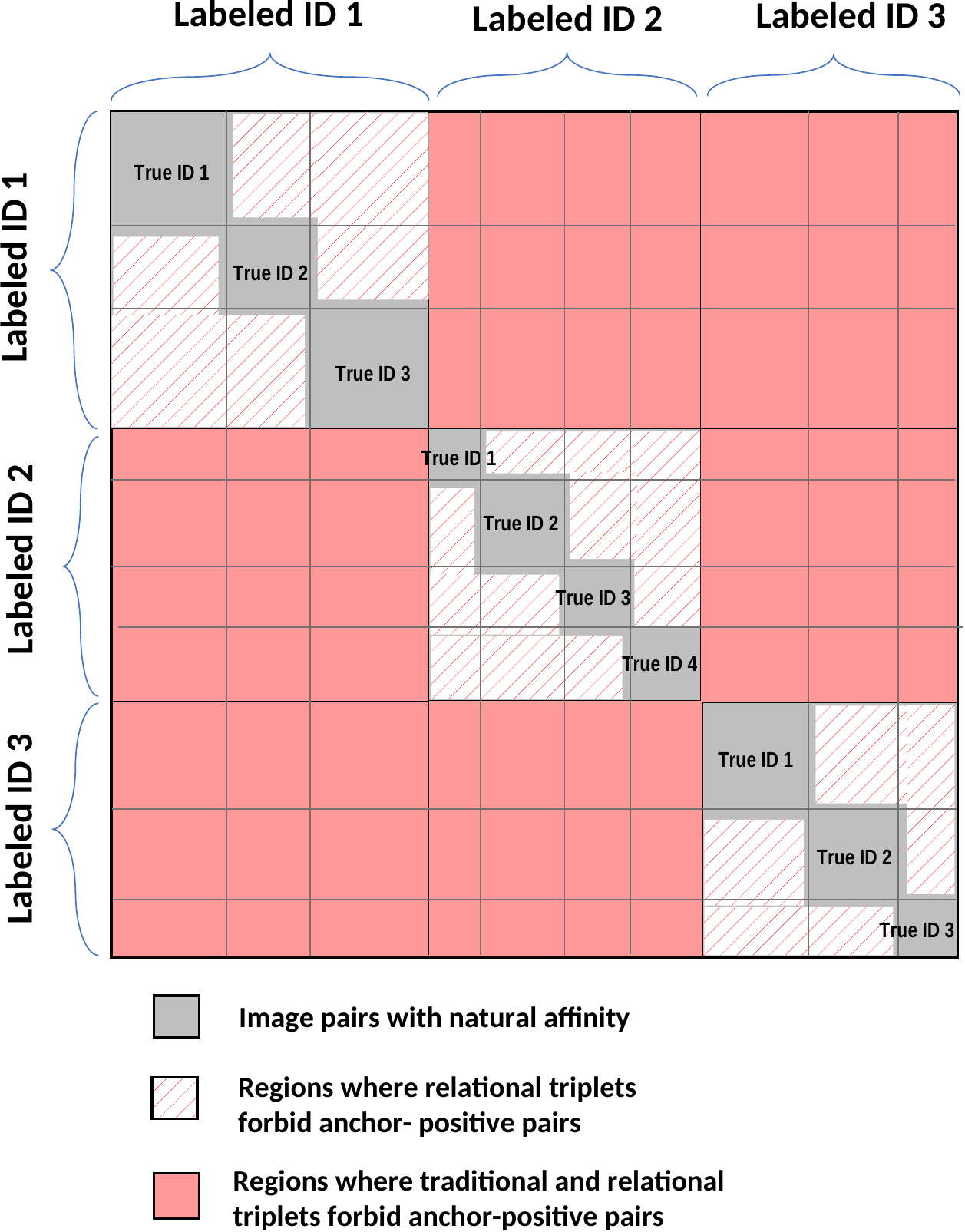}
         \caption{Affinity matrix}
         \label{fig:tri_ideal}
     \end{subfigure}
     \begin{subfigure}[b]{0.5\textwidth}
         \centering
         \includegraphics[width=\textwidth]{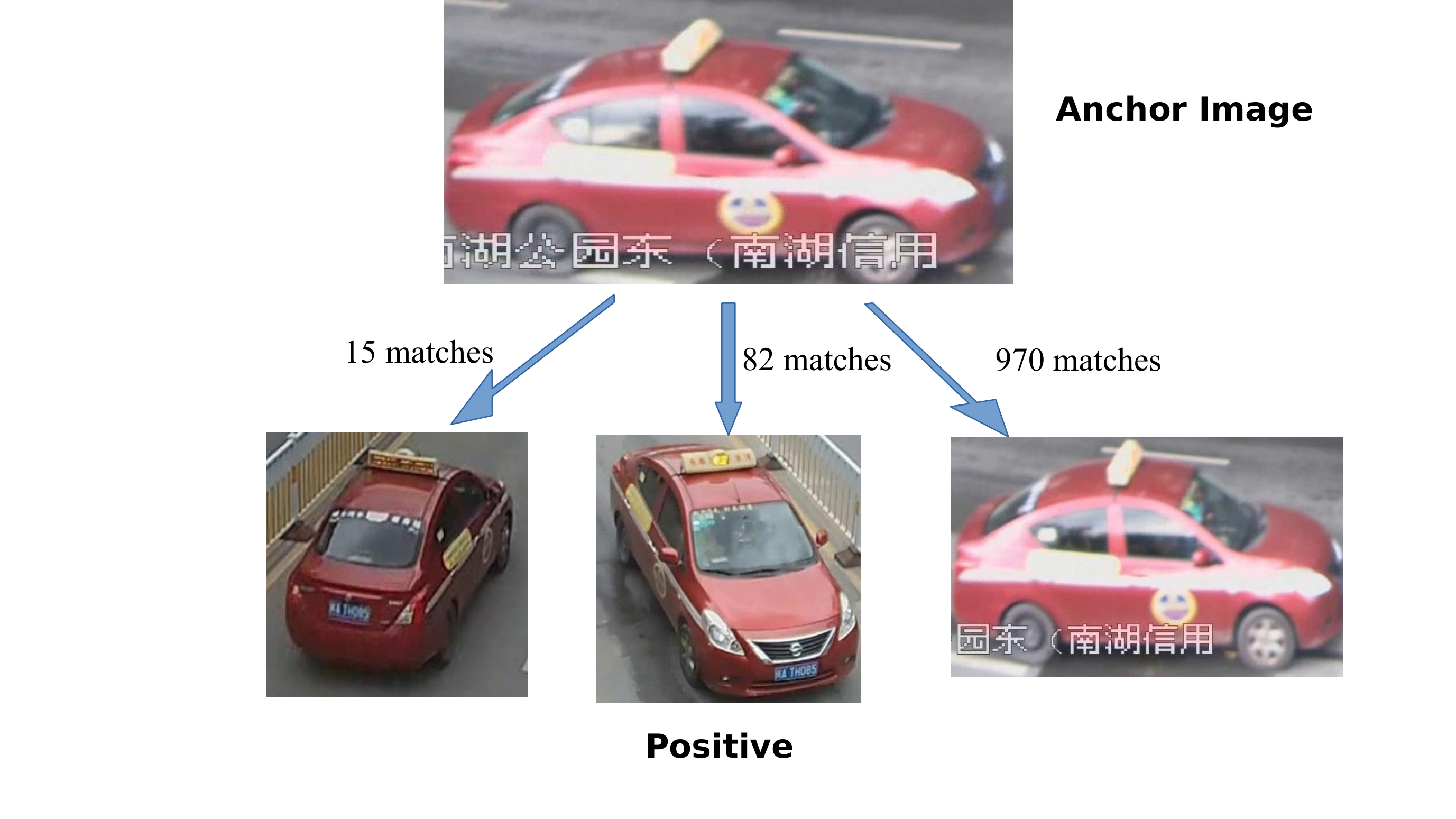}
         \caption{Anchor-positive selection scheme}
         \label{fig:triplet}
     \end{subfigure}
     \caption{\textbf{Representational Schematic for Relation Preserving Triplet Mining}. In figure \ref{fig:tri_ideal}, each ID contains a number of naturally occurring groups. Relational triplets are based on natural groups rather than IDs, thus preventing pathological anchor-positives. In figure \ref{fig:triplet}, observe that the positive shares clear similarities with the anchor(indicating they share a common natural group) but is not a near-duplicate. }
    \label{fig:rptm}
\end{figure*}

\subsection{The Triplet Loss}
\label{sec:tloss}

A triplet loss is defined with respect to three image instances:  Anchor (randomly chosen instance);  Positive (instance that
shares a common  ID with the anchor);  Negative (instance whose ID is different from the anchor).
We denote these instances $\mathbf{x}_a,\mathbf{x}_p$ and $\mathbf{x}_n$, respectively. Given the anchor, positive and negative, the triplet loss is defined as~\cite{schroff2015facenet}:
\begin{equation} \label{eq:trip}
\mathcal{L}_{tri}(\mathbf{x}_a,\mathbf{x}_p, \mathbf{x}_n)  = \max(0, d_{ap} - d_{an} + \alpha),
\end{equation}
where $\alpha$ is the desired margin separation between positive and negative instance,
 $d_{ap} = \|f(\mathbf{x}_a) - f(\mathbf{x}_p)\|$ and $d_{an} = \|f(\mathbf{x}_a) - f(\mathbf{x}_n)\|$. The final triplet-cost is computed by summing the individual triplet losses: 
\begin{equation}
\label{eq:tri_loss}
\quad E_{tri}= \sum^t_{c=1} \mathcal{L}_{tri}(\mathbf{x}_{ac},\mathbf{x}_{pc}, \mathbf{x}_{nc}),
\end{equation}
where $t$ is the total number of triplets. In general, triplet costs are not used in isolation. Instead,  they are combined with the cross-entropy cost from
Eq. \ref{eq:ent_final}, leading to the final cost function:
\begin{equation}
\label{eq:final}
E = \lambda_{ent}E_{ent} + \lambda_{tri} E_{tri},
\end{equation}
where $\lambda_{ent}$ and $\lambda_{tri}$ control the weights given to the cross-entropy loss and triplet-cost respectively.

\begin{figure*}[t]
\begin{center}
   \includegraphics[width=0.7\linewidth]{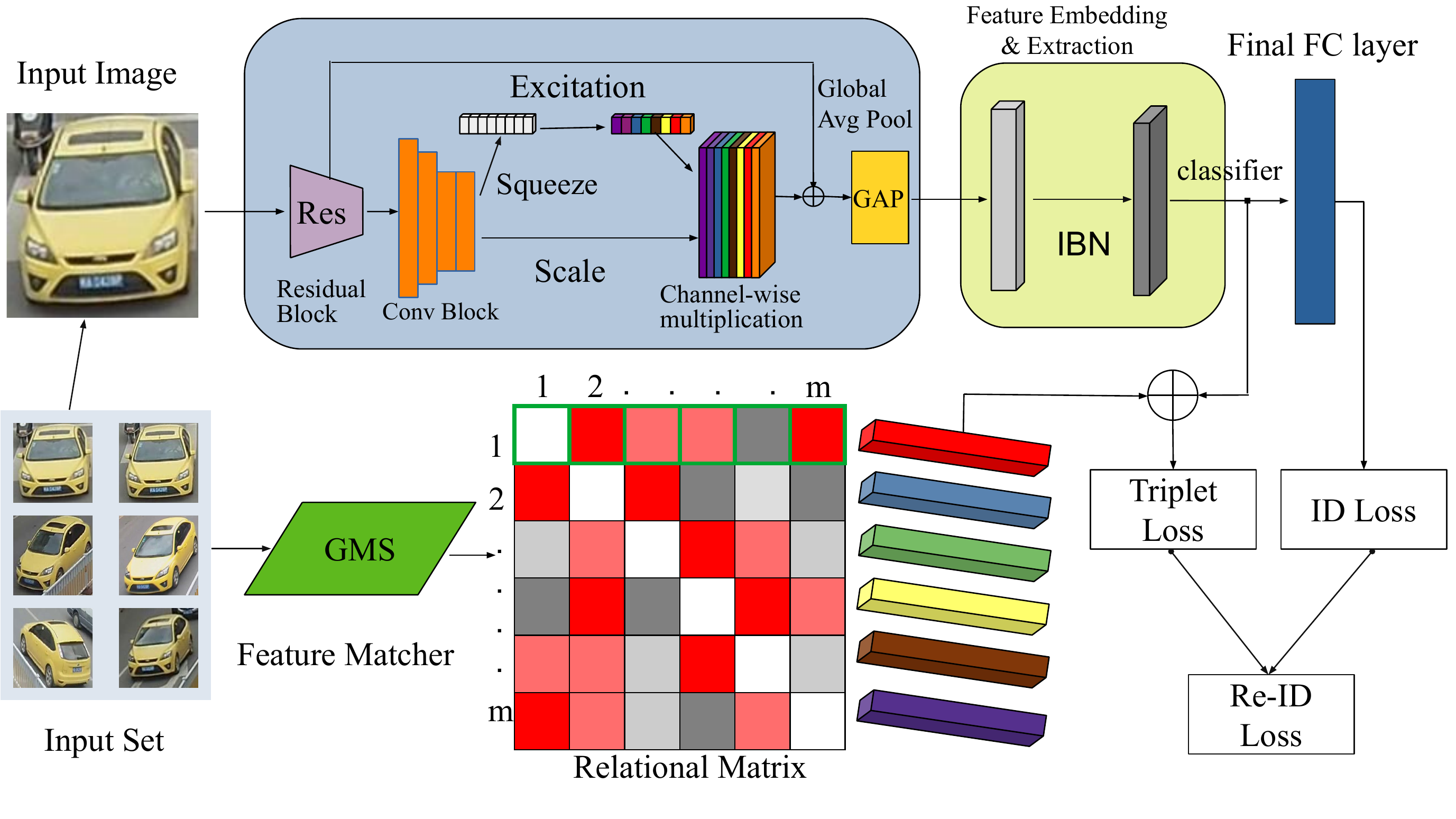}
\end{center}
   \caption{Schematic of a  re-identification network deploying  \textbf{Relation Preserving Triplet Mining}. 
   The RPTM module includes Instance-Batch Normalisation (IBN) and Squeeze-Excitation (SE)  to reduce channel inter-dependencies. The  relational  matrix is estimated using GMS matches and
   is used for triplet selection. }
\label{fig:arch}
\end{figure*}

\section{Relation Preserving  Triplet Mining}
 To prevent training pipelines from stagnating, it is important to implement a good triplet mining scheme. Triplet mining is part of a larger framework
which views features as the key to machine learning. For
example, NetVLAD~\cite{arandjelovic2016netvlad} and many other domain transfer works,
show that adapting features significantly improves performance.
Somewhat similarly, knowledge distillation ~\cite{park2019relational} tries to
compress unwieldy networks into more compact features
for practical deployment. We focus on triplet mining, as most related works in reID use some form of triplet loss and also to effectively highlight \textit{Implicitly Enforced View Consistency}.

Na\"ively incorporating every possible triplet into the loss yields poor results~\cite{hermans2017defense}. 
Instead, training algorithms  employ  triplet-mining, a process which aims to  incorporate
only the most relevant   triplets into the triplet-cost. Unfortunately,   there is no consensus on how relevance can be measured; thus, triplet mining relies on heuristics.
The two most   popular heuristics  are:  hard-negative mining and
semi-hard negative mining. Hard-negative mining focusses on triplets whose negatives are very similar to the anchor. Semi-hard
negative mining shifts the focus from the hardest negatives to negatives close to the decision boundary.
Both heuristics seem sensible and often perform well; however, closer inspection suggests something may be amiss.

Let us perform a thought experiment where we assign the IDs  A and B, to similar   car models. Hard or semi-hard mining
finds the most confusing triplets, leading to the following triplet: front of car A as anchor, rear of car A as positive, and
front of car B as negative. The  triplet is indeed very hard; however, its incorporation into the training cost is  counter-productive.
This is because such a triplet encourages embedding mapping the rear of A to the front of A.
 The embedding  is so counter-intuitive,  it is  unlikely to   generalise well.
 To avoid such pathological cases, we introduce relational triplets, which address
the problem of intraclass separability with greater
attention than other methods.
\subsection{Relational Triplets}
Relational triplets change  the triplet definition  from one based on
 human assigned IDs to one based on naturally occurring groups. 
Formally, we denote the set of training images as $\mathcal{S} = \{\mathbf{x}_1, \mathbf{x}_2, \cdots, \mathbf{x}_{K} \}.$
We hypothesise that these images are members of naturally occurring (and possibly overlapping)  subsets.
The set of subsets is denoted by $\mathcal{N} = \{\mathcal{S}_m\}$,
 where
\begin{equation}
\label{eq:s_natural}
 \mathcal{S} = \bigcup_{\mathcal{S}_{m}\in \mathcal{N}}\mathcal{S}_{m}.
 \end{equation}
 
We use the relational indicator.
 \begin{equation}
 \label{eq:relational}
  C(\mathbf{x}_i, \mathbf{x}_j) =
  	\begin{cases}
    		1,& \text{if } \mathbf{x}_i,  \mathbf{x}_j \text{ share a subset in } \mathcal{N}, \\
    		0,              & \text{otherwise.}
	\end{cases}
 \end{equation}
to denote whether two  instances share  a  natural subset. A  relational triplet is one where 
   the  anchor-positive pair  shares a common natural subset, while the negative does not.
	\begin{equation}   
	\label{eq:rel_tri}
	C(\mathbf{x}_a, \mathbf{x}_p)=1, \quad C(\mathbf{x}_a, \mathbf{x}_n) = 0, \quad C(\mathbf{x}_p, \mathbf{x}_n) = 0. 
	\end{equation}
   
Traditional triplets are a special case of relational triplets, where  
the  given IDs mirror the natural subsets. 
This is not the case in reID,
as we explained in the thought experiment and through the relational diagram in Figure ~\ref{fig:tri_ideal}. Pathological triplets arise when anchor-positive pairs do not have natural affinity (share a common group). Observe that the traditional ID based triplet permits pathological anchor-positive pairs.  In  reID, the natural  subsets likely correspond to object poses. This creates the possibility for  identifying such subsets
 using a  feature matching algorithm. The next section shows how this can be achieved.

%

\subsection{Mining the Relation Preserving Triplets}
\label{sec:tri}

GMS~\cite{bian2017gms} is a modern feature matcher that uses coherence to validate hypothesised feature matches. 
The coherence scheme assumes that a true match hypothesis will be strongly supported by many other match hypotheses between
neighbouring region pairs, while a false
  match hypothesis will not.  The coherence-based   validation is notably better than the  traditional ratio test~\cite{lowe1999object}.
  This allows GMS to reliably match features across significant viewpoint changes while simultaneously ensuring  
  few  matches between image pairs with nothing in  common. 
  As a result,  the presence of GMS matches between image pairs provides a good approximation of the  
    relational indicator in Eq. \ref{eq:s_natural}. GMS is quite effective in reID systems to quantify the innate relation between images and is crucial in establishing implicitly enforced view consistencies.

    While GMS has few errors, errors do occur. To ensure
    an anchor-positive pair has a relational indicator of one, 
    we set  the positive instance
of each anchor to be the image instance
whose number of GMS matches with the anchor is closest to the threshold $\tau$. Here, we accept that setting
similar anchor-positive pairs leads to poorer training. Hence,
we use a middle-ground approach for anchor-positive selection,
which we call $RPTM_{mean}$, in which
$\tau$ is set as the average number of GMS matches in the set of nonzero pairwise GMS matches between the anchor
and all other images. More formally, 
for two images, $\mathbf{x}_i, \mathbf{x}_j$ we predict that the natural relational indicator is true, $C(\mathbf{x}_i, \mathbf{x}_j) = 1$, if the number matches between them exceed $\tau$.

The above provides a semi-hard positive mining, that ensures anchor-positive pairs satisfy the relational indicator in Eq. \ref{eq:s_natural},
while also ensuring that the positive differs significantly from the anchor. An example is shown in Figure \ref{fig:triplet}.We define negatives using batch hard-triplet mining~\cite{hermans2017defense}. If $\mathcal{S}_{b} = \{\mathbf{x}_j\}$
denotes the set of instances in a batch that do not share an ID with $\mathbf{x}_a$,
the negative is
\begin{equation}
\mathbf{x}_n = \argmin_{\mathbf{x}_j \in \mathcal{S}_{b} } \left(\|(f(\mathbf{x}_{a})- f(\mathbf{x}_{j})\|\right).
\end{equation}

Observe that the triplets defined in this manner satisfy Eq.~\ref{eq:rel_tri}, making them  
relation preserving triplets. Given such triplets, the final embedding can be obtained by 
minimising the cost function in Eq.~\ref{eq:final}. As evidenced by the mining strategy, RPTM allows for an intrinsic understanding
of the viewpoint and pose without hard coded pose estimation.

\section{Implementation Details}

 A schematic of the network architecture is provided in Figure~\ref{fig:arch}. In this section we discuss the model layout, elaborating on the comparative feature matching pipeline in Section \ref{feat-match} and the model structure with RPTM in Section \ref{net}. To test and highlight the universality of RPTM and its ability to generalise the training pipeline due to its novel triplet mining scheme, we put limitations on network and parameter tuning across all datasets.
 
\begin{figure}[t]
\centering
    \begin{subfigure}{1.0\linewidth}
\includegraphics[trim=0 140 110 0,clip, width=1.0\linewidth]{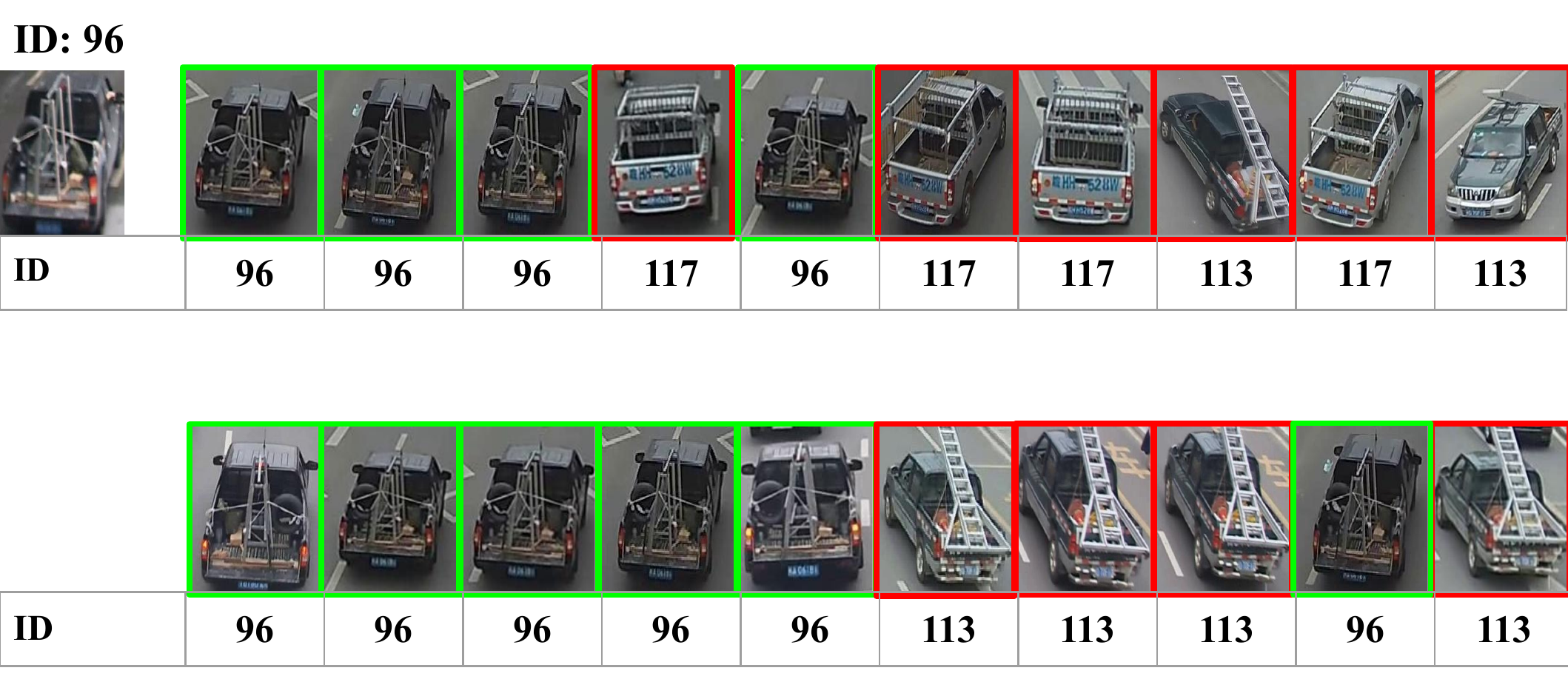} 
\put(-250, 70) {R1}
\put(-250, 15) {R2}

\label{fig:4a}
    \end{subfigure}\hfill
    \par\bigskip
    \begin{subfigure}{1.0\linewidth}
\includegraphics[trim=0 140 110 0,clip,width=1.0\linewidth]{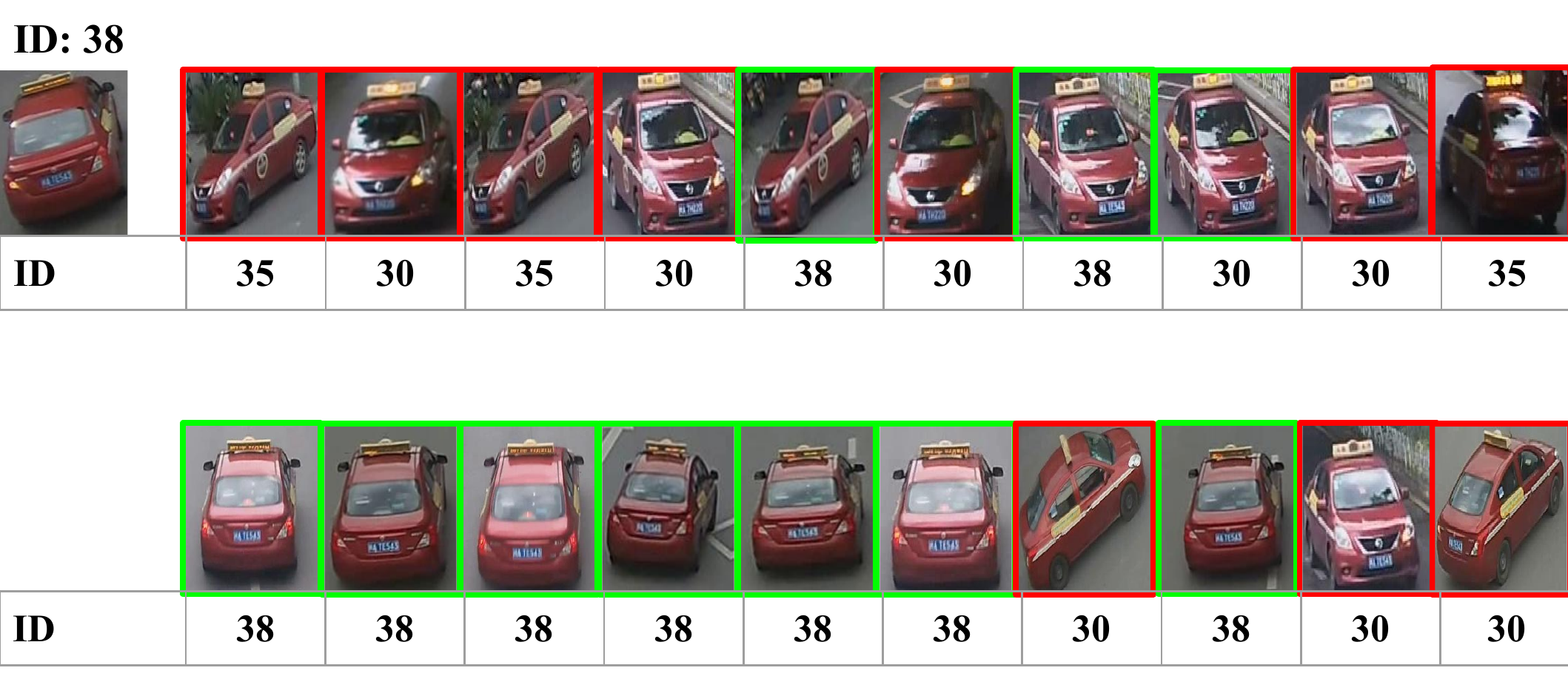}
\put(-250, 70) {R1}
\put(-250, 15) {R2}
\label{fig:4b}
    \end{subfigure}

\caption{
Qualitative retrieval results for bad targets without RPTM(R1) and with RPTM(R2). Correct identifications are outlined in green; wrong ones are outlined in red.  RPTM clearly aligns backbone models with better pose awareness and provides fine-grained attention.
}
    \label{fig:topk}
\end{figure}

\subsection{Feature Matching}
\label{feat-match}
As discussed before, we use GMS feature matching to guide our triplet-mining process, in order to implement semi-hard positive mining.  In theory,
we need to establish GMS matches between an anchor and every other image in the dataset. In practise, we use image IDs as guides
to the natural groupings and restrict the matching to only images that share a common ID with the anchor. This greatly reduces computational cost in triplet mining. Feature matching is performed on  images that have been resized to (224, 224).  The GMS
feature matching parameters are: 10,000 ORB features whose orientation parameter is set to true and
nearest neighbours are identified with the brute-force hamming distance. All other parameters are set according to the guidelines reported by ~\cite{bian2017gms}. 
After matching, the number of matches between image pairs is stored in a relational matrix $m \times m$, where $m$ is the number of training images.

\subsection{Neural Network}
\label{net}
For fair comparison of our results with established benchmarks, we chose ResNet-50 and ResNet-101 pretrained on ImageNet as our backbone. Our RPTM module includes instance-batch-normalization and a squeeze-excitation layer\cite{hu2018squeeze}.  
The weights of this network are trained
by minimising the loss function in Eq. \ref{eq:final}. This network is trained using triplets defined through our Relation Preserving Triplet Mining (RPTM) in Section \ref{sec:tri}. 

The images are resized to (240,240) for vehicle reID and (300,150) for person reID.
Data augmentation is applied,  with random flipping, random padding, random erasing and colour jitter (randomly changing contrast, brightness, hue and saturation) all  activated.  
 Stochastic Gradient Descent(SGD) is used as the optimiser for the model. The initial learning rate is initialised at 0.005 and is  set to  decay by a factor of 0.1 every 20 epochs.
 The model is trained for 80 epochs with a batch size of 24.
 Training parameters are fixed for all datasets.~\footnote{
 These 
parameters are significantly less computationally demanding  than those used by recent state-of-the-art models~\cite{Fu_2022_CVPR,he2021transreid,rao2021counterfactual,wang2022nformer,zhu2022dual}}.
Finally,  Figure~\ref{fig:topk} provides qualitative comparisons showing that RPTM's top-k-ranked retrievals are significantly better than its backbone network (we showcase top-k results alternatively (top-1, top-3...top-19). We focus on demonstrating the quality of gallery image retrieval for query samples by RPTM by comparing top-k retrieval results with and without the RPTM pipeline.


\begin{table*}[h]
\small
\begin{center}
\centering
\begin{tabular}{P{3.4cm}|P{1cm}P{1cm}P{1cm} | P{1cm}P{1cm}P{1cm} | P{1cm}P{1cm}P{1cm}  }
 \hline
Model & \multicolumn{3}{l|}{Small (query size=800)} & \multicolumn{3}{l|}{Medium (query size=1600)} & \multicolumn{3}{l}{Large (query size=2400)}   \\
 & mAP & r=1 &r=5 & mAP & r=1 &r=5& mAP & r=1 &r=5\\
 \hline
 \hline
 C2F-Rank~\cite{guo2018learning} &63.50& 61.10 &81.70 &60.00 &56.20 &76.20 &53.00 &51.40 &72.20\\
 AGNet~\cite{zheng2019attributes} &76.06& 73.14& 86.25 &73.39 &70.77 &81.75 &71.75& 69.10& 80.40\\
 ANet~\cite{quispe2021attributenet}&-& 86.00 &97.40&-& 81.90 &95.10 &-&79.60 &92.70\\
 VANet~\cite{chu2019vehicle} & -& 88.12 &97.29& -& 83.10 &95.14 &- & 80.35 &92.97\\
 Smooth-AP~\cite{brown2020smooth} & -& 94.90 &\textbf{97.60}& -& \textbf{93.30} &\textbf{96.40} &- & 91.90 &\textbf{96.20}\\
 \hline
 RPTM (ResNet-50) & \textbf{82.30} & \textbf{95.00} & 96.70& \textbf{79.90} & 92.50 & 96.20 & \textbf{78.60}&\textbf{92.10}&95.70\\
  \hline
 
 QD-DLP~\cite{zhu2019vehicle} &76.54 &72.32 &92.48& 74.63& 70.66& 88.90& 68.41& 64.14 &83.37\\
 AAVER~\cite{khorramshahi2019dual} &-&74.69& 93.82 &-&68.62 &89.95&- &63.54 &85.64\\
 VehicleNet~\cite{zheng2020vehiclenet}&-&83.64 &96.86&-&81.35 &93.61 &-&79.46 &92.04\\
 \hline
RPTM (ResNet-101) & \textbf{84.80} & \textbf{95.50} & \textbf{97.40}& \textbf{81.20} & \textbf{93.30} & \textbf{96.50} & \textbf{80.50}&\textbf{92.90}&\textbf{96.30}\\

 

\hline

\end{tabular}

\end{center}
\caption{Comparison with state-of-the-art methods on VehicleID. RPTM provides the best retrieval results in all three test sets, with notably better performance in the large test set.}
\label{tab:vid}
\end{table*}

\begin{table}[h]
\small
\begin{center}
\centering
\begin{tabular}{P{3.4cm}|P{1cm}|P{1cm}|P{1cm}  }
 \hline

 \hline
 Model& mAP &r = 1&r = 5\\
 \hline
 \hline
 
 SPAN~\cite{chen2020orientation}& 68.90	& 94.00	& 97.60\\
 PAMTRI~\cite{tang2019pamtri} & 71.88	& 92.86	& 96.97\\
 PVEN~\cite{meng2020parsing} & 79.50 & 95.60 & 98.40\\
 TBE~\cite{sun2021tbe} & 79.50 & 96.00 & \textbf{98.50}\\
 \hline
 RPTM (ResNet-50) & \textbf{79.90} & \textbf{96.10} & \textbf{98.50}\\
 \hline
 GAN+LSRO$^{*}$~\cite{wu2018joint} & 64.78 &88.62 & 94.52\\
 SAVER$^{*}$~\cite{khorramshahi2020devil} & 82.00 & \textbf{96.90} & 97.70\\
\hline
 RPTM (ResNet-50) $^{*}$ & \textbf{86.40} & 96.70 & \textbf{98.00}\\
 \hline

CAL~\cite{rao2021counterfactual} & 74.30 & 95.40 & 97.90\\
TransReID~\cite{he2021transreid} & 80.60 &\textbf{96.80} & --\\
 \hline
 RPTM (ResNet-101) & \textbf{80.80} & 96.60 & \textbf{98.90}\\
 \hline
AAVER$^{*}$~\cite{khorramshahi2019dual} &66.35&	90.17&	94.34\\
DMT$^{*}$~\cite{he2020multi} & 82.00 & 96.90 & --\\
VehicleNet$^{*}$~\cite{zheng2020vehiclenet} & 83.41 & 96.78 & --\\
Strong Baseline$^{*}$~\cite{huynh2021strong} & 87.10 & 97.00 & --\\
\hline
 RPTM (ResNet-101) $^{*}$ & \textbf{88.00} & \textbf{97.30} & \textbf{98.40}\\
 \hline


\end{tabular}
\end{center}

\caption{Comparison with the state-of-the-art results on the Veri-776 dataset. The $*$ indicates the usage of re-ranking.}
\label{tab:veri}
\end{table}

\section{Experiments}
\subsection{Datasets}
\textbf{VehicleID}~\cite{liu2016deep} 
allows us to  test RPTM's scalability by offering multiple,  progressively larger (and  harder) test-sets. We evaluate our algorithm with 800, 1600 and 2400 labels for testing.
\textbf{Veri-776}~\cite{liu2016large} is a widely used benchmark with a diverse range of viewpoints for each vehicle and is designed to provide more constrained but highly realistic conditions.  \textbf{DukeMTMC}~\cite{ristani2016performance} is a person re-identification benchmark with 1,404 distinct classes. While our focus is vehicle reID, we include this benchmark to show our algorithm
can generalise to other problems.

\subsection{Evaluation Metrics}

\begin{table}[t]
\begin{center}
\small
\begin{tabular}{P{3.4cm}|P{1cm}|P{1cm}|P{1cm}  }
\hline

\hline
 Model& mAP &r = 1&r = 5\\
 \hline

 P2-Net~\cite{guo2019beyond} &73.10&	86.50&	93.10\\
 GPS~\cite{nguyen2021graph} &78.70&	88.20&	95.20\\
 PNL~\cite{Fu_2022_CVPR} &79.00 & 89.20&--\\
 SCSN~\cite{chen2020salience} &79.00 & 91.00&--\\
 \hline
RPTM (ResNet-50) & \textbf{80.20} & \textbf{91.40} & \textbf{95.80}\\
 \hline
 Top-DB-Net$^{*}$~\cite{quispe2020top} & 88.60	& 90.90 &--\\
 NFormer$^{*}$~\cite{wang2022nformer} & 83.40 & 89.50 & --\\
 st-reID$^{*}$~\cite{wang2019spatial} & \textbf{92.70} &\textbf{94.50}&  \textbf{96.80}\\
 \hline
 RPTM (ResNet-50)* & 87.50 & 92.30 & 95.20\\
 \hline
 
 PAT$^{*}$~\cite{li2021diverse} & 78.20 & 88.80 & --\\
 LDS$^{*}$~\cite{zang2021learning} & \textbf{91.00}& 92.90&--\\	
 \hline
RPTM (ResNet-101)* & 89.20 & \textbf{93.50} & \textbf{96.10}\\
 \hline
 
\hline
\end{tabular}
\end{center}
\caption{Comparison on the DukeMTMC benchmark. RPTM provides competitive results even though it is not tuned for person reID.  $*$ indicates re-ranking.}
\label{tab:duke}
\end{table}

Rankings are scored according to  the  protocols suggested in~\cite{liu2016deep, liu2016large} and all methods are reported with mean average precision (mAP) and Cumulative Matching Characteristics(CMC). 
For the  Veri-776 and DukeMTMC datasets, we also use re-ranking~\cite{zhong2017re}, which refines the final rankings by considering the k-reciprocal nearest-neighbours
of both the query and retrieved images, effectively improving upon the pairwise distance result that is used to quantify mAP and top-k ranking accuracies. Re-ranking  is not adopted for VehicleID because
there is often only one true match ID in the gallery set~\cite{khorramshahi2019dual}. We split past works based on the complexity of the backbone network, with our results on ResNet-50 and ResNet-101 backbones.

\subsection{Comparison with State-of-the-art}

\textbf{VehicleID:}
Table \ref{tab:vid} shows that RPTM achieves state-of-the-art results on the challenging VehicleID dataset, indicating RPTM's scalability across vehicle datasets. Although not exceeding Smooth-AP\cite{brown2020smooth}, table \ref{tab:smoothap} shows a drop in performance by Smooth-AP on Veri-776 and DukeMTMC.

\textbf{Veri-776:}
As shown in table \ref{tab:veri}, RPTM surpasses the recent state-of-the-art vehicle reID models.
These results are very respectable, especially if we consider the fact that well-performing algorithms like  VehicleNet~\cite{zheng2020vehiclenet} uses  supplementary data for training. We also edge out Strong Baseline~\cite{huynh2021strong}, which uses deeper backbones and larger images. In addition, RPTM's
training scheme is very simple, as it only requires gradient descent on a well-defined loss.

\textbf{DukeMTMC:}Table \ref{tab:duke} shows  RPTM achieves competitive results at person reID, despite training parameters tuned to vehicle datasets. 
With the exception of changing the image size
to account for the aspect ratio of input images, no changes were made to the RPTM network or training parameters. These results are respectable for a network whose training parameters are tuned  for a different  task.

\textbf{Discussion:} 
Table \ref{tab:vid},  \ref{tab:veri} and \ref{tab:duke}, show that incorporating 
RPTM to feature learning techniques make them more   effective at re-identification. Performance improvements are especially
notable on more difficult datasets like VehicleID and harsher evaluation metrics (mAP). These performances are quite remarkable when we take into account that RPTM uses constant training parameters for all three datasets. Most deep-learning
algorithms require parameters to be tweaked from dataset to dataset, and RPTM's capability in this respect is an indication
that relational aware triplet choice makes the triplet losses better conditioned.

To demonstrate the challenge of maintaining constant training parameters, we trained Smooth-AP~\cite{brown2020smooth} 
 on  two other datasets, while using the training parameters of Table~\ref{tab:vid}, as shown in Table~\ref{tab:smoothap}. We also acknowledge the use of Visual Transformers (ViT) in TransReID by He \textit{et al.}~\cite{he2021transreid}, demonstrating impressive results, albeit using camera embeddings and viewpoint labelling. Although RPTM uses universal parameters that are compliant to low compute requirements, we still achieve state-of-the-art results compared to transformer-based ReID models. As an additional experiment, using the increased parameter settings defined in TransReID, we further improve our retrieval results, achieving an mAP of \textbf{82.5\%(w/o re-ranking)} on Veri-776.
 
 \begin{table}[h]

\begin{center}
\begin{tabular}{c|c|c|c  }
 Method & mAP &r = 1&r = 5\\
 \hline
 \hline
Smooth-AP (Veri-776) & 79.40& 91.10&94.20\\
RPTM (Veri-776) & \textbf{88.00} & \textbf{97.30}&\textbf{98.40}\\

 \hline
 Smooth-AP (DukeMTMC) & 65.70& 79.90&88.40\\
 RPTM (DukeMTMC) & \textbf{89.20} & \textbf{93.50} & \textbf{96.10}\\
 \hline
\end{tabular}
\end{center}
\caption{Performance of Smooth-AP~\cite{brown2020smooth} (ResNet-101 backbone) on Veri-776 and DukeMTMC, with re-ranking.}
\label{tab:smoothap}
\end{table}

\subsection{Ablation Study}
\textbf{Image Size.} We begin by investigating how image size impacts re-identification.  Table~\ref{tab:Img}
shows that the evaluation metrics improve as the image size increases, a finding that is mirrored by many other reID algorithms,
which often seek to use the largest possible image. However, we find that performance peaks at (240, 240) on Veri-776 and VehicleID, which validates RPTM's ability to achieve state-of-the-art results at lower resolutions compared to other benchmarks.

\begin{table}[t]

    \begin{subtable}[h]{0.45\textwidth}
        \centering
        \begin{tabular}{c|cc|cc}
    \hline
    Model& \multicolumn{2}{l|}{Veri-776} & \multicolumn{2}{l}{VehicleID(small)}    \\
    & mAP & r=1 & mAP & r=1 \\
    \hline
    $RPTM_{128\times128}$ &56.5 & 84.5& 72.5&89.0\\
    $RPTM_{160\times160}$ & 74.8& 92.4& 80.5&91.8\\
    $RPTM_{224\times224}$ & 85.1& 95.2&83.1& 92.9\\
    $RPTM_{240\times240}$&  \textbf{88.0} & \textbf{97.3}&\textbf{84.8}&\textbf{95.5}\\

     \hline
       \end{tabular}
       \caption{Image size ablation}
    \label{tab:Img}
    \end{subtable}
    \hfill
    \begin{subtable}[h]{0.45\textwidth}
        \centering
        \begin{tabular}{c|cc|cc}
    \hline
     Model& \multicolumn{2}{l|}{Veri-776} & \multicolumn{2}{l}{VehicleID(small)}    \\
    & mAP & r=1 & mAP & r=1 \\
    \hline

    $RPTM_{min}$ & 86.3 & 95.9 & 82.1&93.9\\
    $RPTM_{mean}$   &  \textbf{88.0} & \textbf{97.3} & \textbf{84.8}& \textbf{95.5}\\
    $RPTM_{max}$ & 82.2 & 95.6 & 79.8&93.1\\
    \hline
        \end{tabular}
        \caption{Thresholding ablation}
        \label{tab:thresh}
     \end{subtable}
     \caption{(a) ReID performance with increasing image size. mAP and rank-1 increase
with image size until (240, 240), after which performance plateaus. (b) Comparing   positive selection thresholds. $RPTM_{min, mean, max}$ correspond to hard positive, semi-hard  positive and easy positive-mining.}
     \label{tab:ablation}
\end{table}

\textbf{Threshold for Positive Selection}
 Section \ref{sec:tri} suggests positive images are chosen using  a threshold, $\tau$, which  is the mean number
of non-zero matching results. We denote this  scheme  $RPTM_{mean}$ (semi-hard positive mining). There are a number of  alternatives. One possibility is to fix $\tau$ on a low number of matches, such as $10$. We term this scheme $RPTM_{min}$.  
The scheme ensures anchor-positive pairs are not  near duplicates  
and  corresponds to hard positive mining.    The drawback is a vulnerability to occasional matching errors. Another possibility is to set $\tau$ to the largest number of matches that the anchor image has. We term this  
 $RPTM_{max}$. This eliminates any vulnerability to GMS matching errors but
 sacrifices the  positive image's distinctiveness. This corresponds to easy positive mining.   Table~\ref{tab:thresh} indicates that $RPTM_{mean}$ has the best performance; hence, it is adopted as our default mining scheme.

\section{Conclusion}

In this work, we have shown that respecting natural data groupings within classes can help significantly improve triplet mining, not only facilitating the selection of better anchor-positive pairs but also, consequently, creating a more tractable optimisation procedure that leads to better generalisation. 
To that end, we introduced Relation Preserving Triplet Mining (RPTM), a triplet-alignment scheme to generate samples wary of the inverse-variability problem, proving that implicitly enforced view consistencies can significantly improve the reID pipeline. 
We showed how feature matches could be used to develop relation-aware triplet mining, leading to a  better conditioned  triplet loss, creating feature learners with enhanced training stability.
Moreover, we highlighted that RPTM outperforms recent reID models while maintaining constant training parameters across datasets.
Finally, we believe our research can be extended to Unsupervised Domain Adaptation for even better scalabilty across reID datasets due to RPTM's implicit ability to align similar features, Anomaly Detection in reID to see how noise affects performance, and also for general deep image retrieval to evaluate RPTM outside re-identification.

{\small
\bibliographystyle{ieee_fullname}
\bibliography{egbib}
}

\end{document}


\title{Relation Preserving Triplet Mining for Stabilizing the Triplet Loss in Re-identification Systems: Supplementary Material}


\author[1,2]{Adhiraj Ghosh}
\author[1,3]{Kuruparan Shanmugalingam}
\author[1]{Wen-Yan Lin}
\affil[1]{Singapore Management University}
\affil[2]{University of Tübingen}
\affil[3]{University of New South Wales}
\maketitle
\thispagestyle{empty}

\section*{\centering{Overview}}
In this document, we provide additional results and technical details. Firstly, Section \ref{arch} provides a detailed description of the network architecture. Section \ref{out} analyses the presence of outliers in reID datasets and how they degrade feature learning. Section \ref{tr} provides qualitative analysis of the triplet mining strategy used by RPTM in semi-hard positive sampling. Section \ref{rr} demonstrates the role played by re-ranking in improving evaluation results. Finally, we show more visual comparisons of our model against current state-of-the-art results in Section \ref{exp}.  

\section{Architecture Details and Hyperparameters}
\label{arch}
A major focus of our research was to create a universal model that performs well across benchmarks, for both vehicle re-id and person re-id datasets. As mentioned in Section 5.2 of the main paper,we observed that RPTM performs well on the ResNet backbone with Squeeze-Excitation~\cite{hu2018squeeze} and an Instance Batch Normalisation (IBN) appendage. Table \ref{tab:hyp} provides the details of the hyperparameters of the universal network that generated state-of-the-art results in vehicle re-id and comparable results in person re-id.

\begin{table}[h]
\begin{center}
\begin{tabular}{l|c}
 & RPTM (ResNet101 Baseline)\\
 \hline
 Input Size(Veri-776) & $240\times240$\\
 Input Size(VehicleID) & $240\times240$\\
 Input Size(DukeMTMC) & $300\times150$\\
 Train Batch Size & 24\\
 Test Batch Size & 100\\
 Workers & 8\\
 Optimizer & SGD\\
 Momentum & 0.9\\
 Weight Decay & $5e^{-4}$\\
 Learning Rate & 0.005\\
 Scheduler & MultiStepLR\\
 Decay Factor & 0.1\\
 Margin (triplet loss) &0.3\\ 
 $\lambda_{tri}$ & 2\\
 $\lambda_{xent}$ & 0.5\\
 Stride &1\\
 Droprate & 0.2\\
 Pooling & Average\\
 Pre-trained &ImageNet\\
 Feature Dimension & 2048\\
\end{tabular}
\end{center}
\caption{Hyperparameters of RPTM (ResNet101 Baseline) used for implementing RPTM on reID benchmark datasets.}
\label{tab:hyp}
\end{table}

\begin{figure}[h]
\begin{center}
   \includegraphics[width=1.0\linewidth]{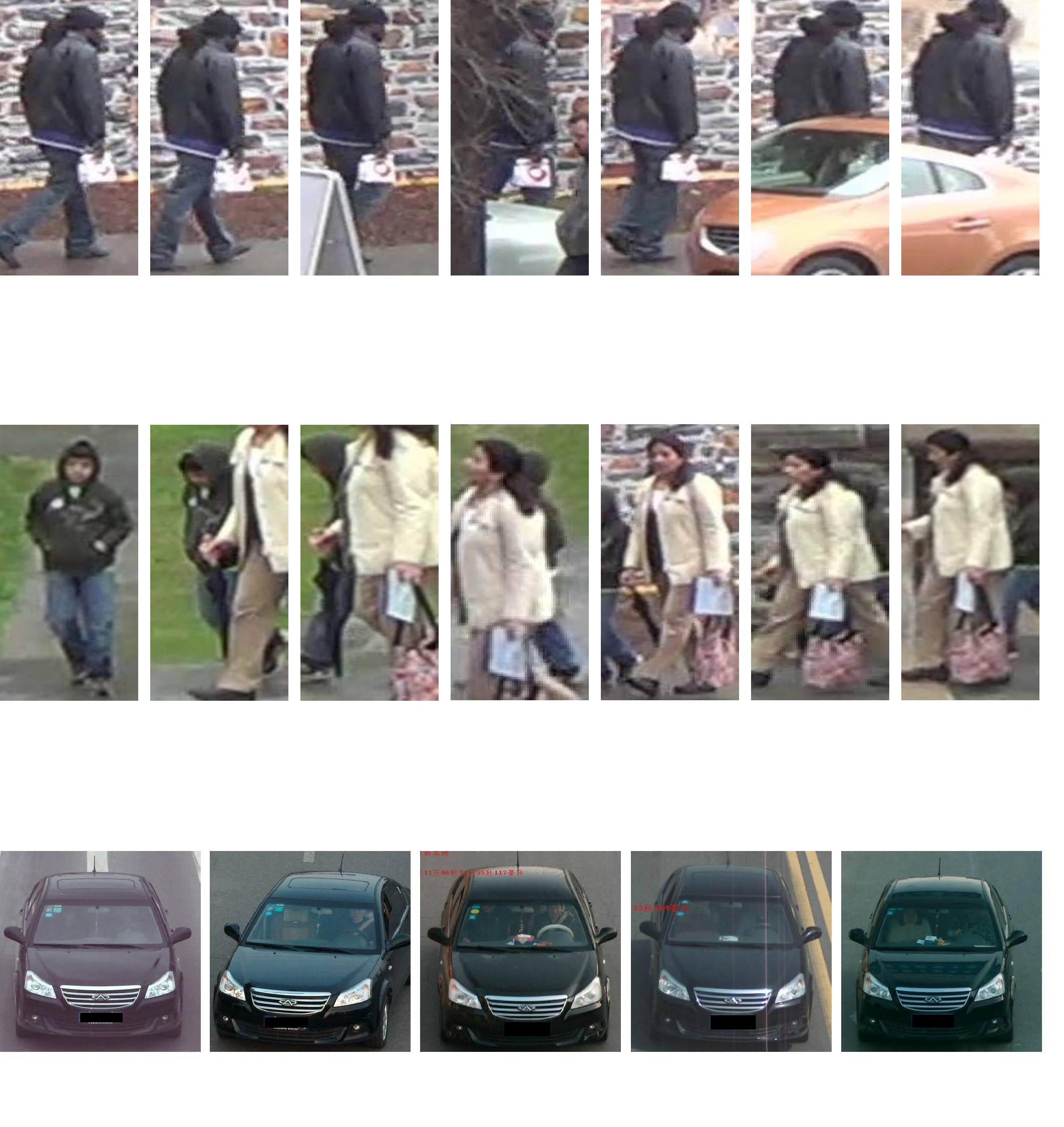}
   \put(-235, 255){ID:70}
   \put(-200, 255){ID:70}
   \put(-167, 255){ID:70}
   \put(-132, 255){ID:70}
   \put(-99, 255){ID:70}
   \put(-67, 255){ID:70}
   \put(-35, 255){ID:70}
   \put(-235, 160){ID:13}
   \put(-200, 160){ID:13}
   \put(-167, 160){ID:13}
   \put(-132, 160){ID:13}
   \put(-99, 160){ID:14}
   \put(-67, 160){ID:14}
   \put(-32, 160){ID:14}
   \put(-232, 65){ID:2395}
   \put(-185, 65){ID:2395}
   \put(-138, 65){ID:3004}
   \put(-92, 65){ID:3392}
   \put(-45, 65){ID:6478}
\end{center}
   \caption{Outliers serve as a hindrance to proper training of reID models. Standard models are unable to focus on fine-grained details and resolve outlier cases during training. Some examples of outliers in the DukeMTMC(top two rows) and VehicleID(bottom row) datasets.  The first row shows the effect of occlusions. The second row shows an overlap between object tracklets of two persons. The last row shows the distribution of the same vehicle model with the same colour across several IDs.}
\label{fig:prob}
\end{figure}

\begin{figure}[t]

\centering
    \begin{subfigure}{1.0\linewidth}
\includegraphics[trim=50 0 0 0,clip, width=1.0\linewidth]{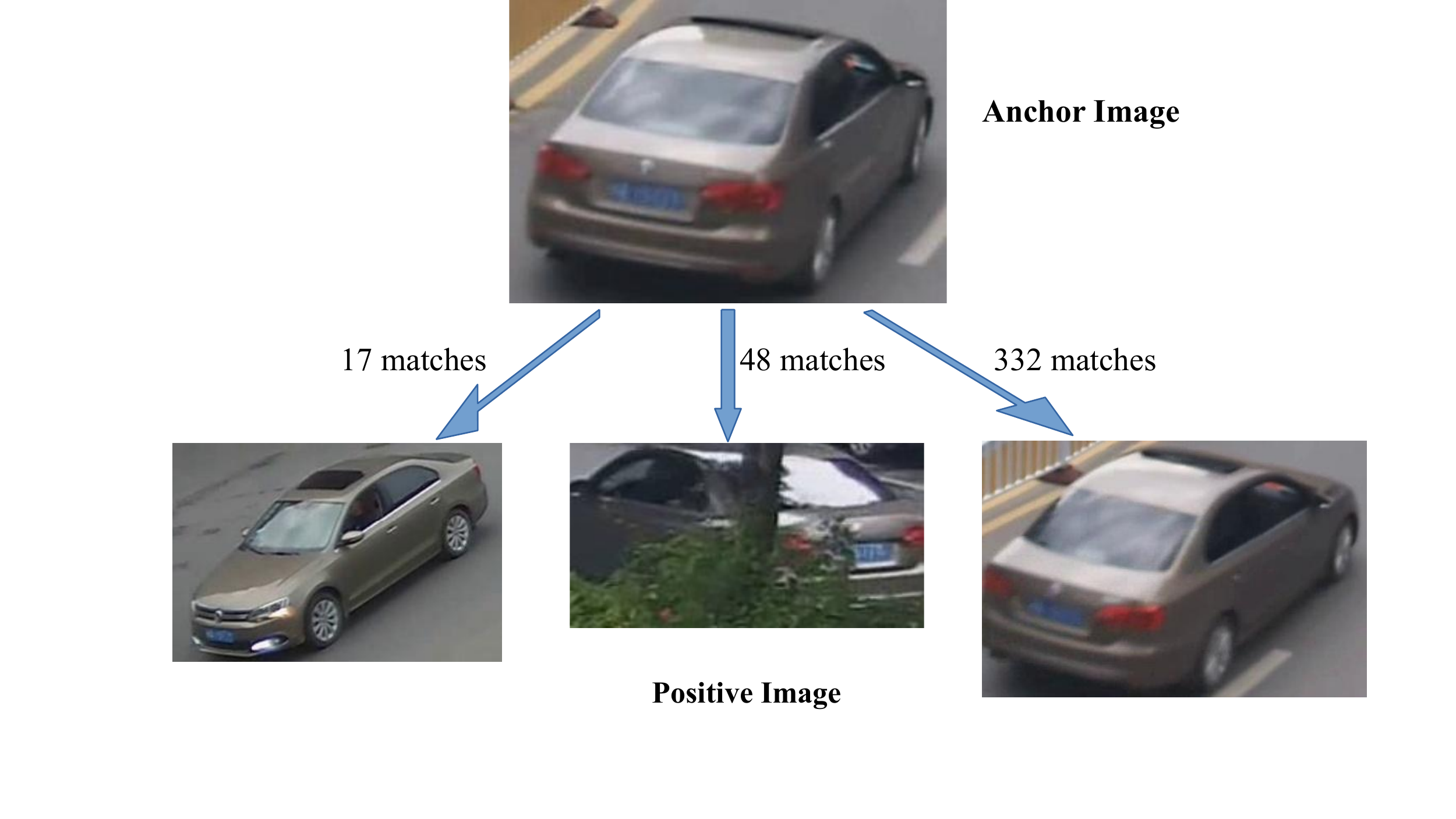} 
\put(-120, 100) {}
\caption{Anchor-Positive Pair Selection by RPTM: Veri-776}
\label{fig:4a}
    \end{subfigure}\hfill
    \par\bigskip
    \begin{subfigure}{1.0\linewidth}
\includegraphics[trim=90 0 50 0,clip, width=1.0\linewidth]{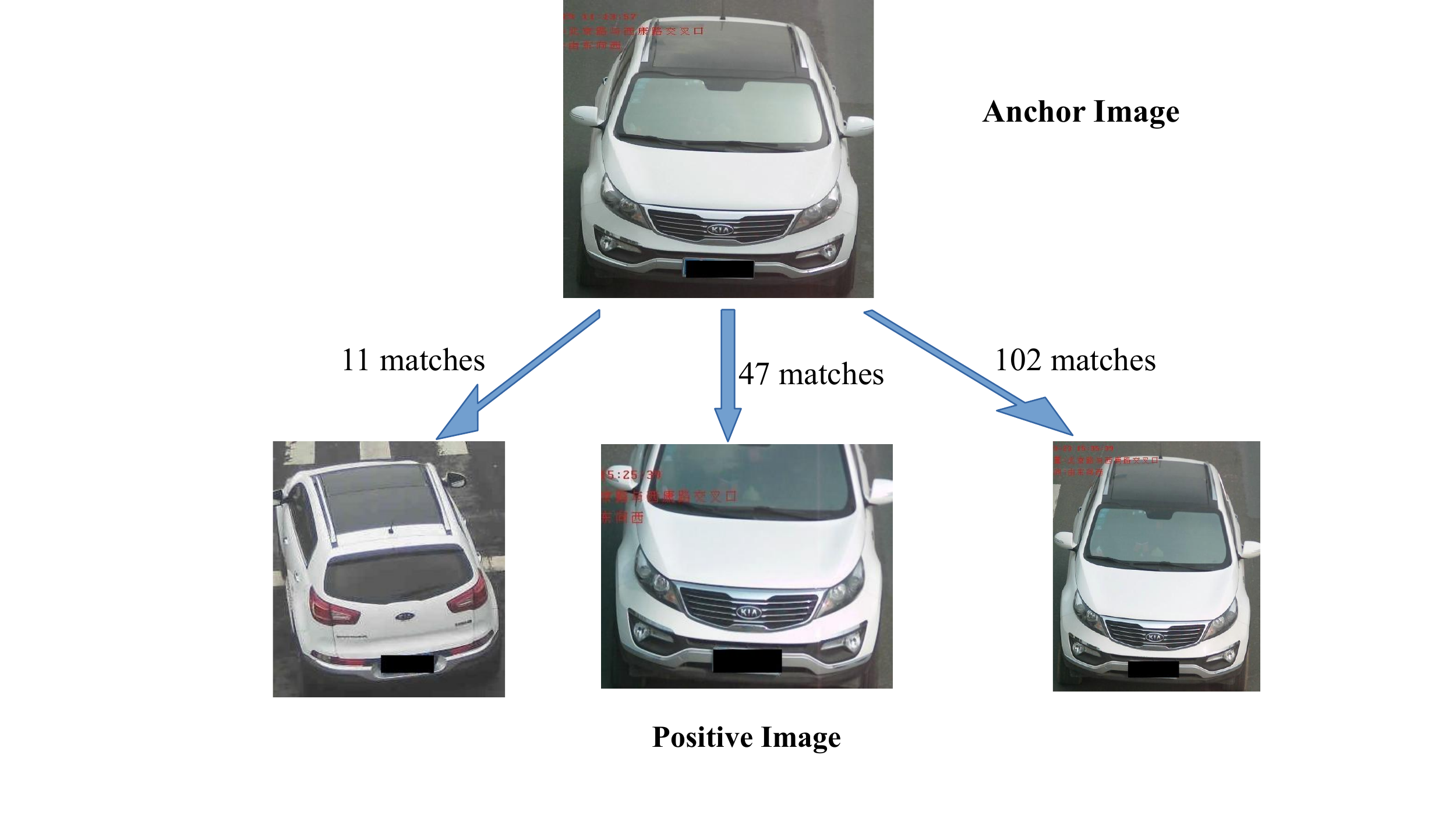} 
\caption{Anchor-Positive Pair Selection by RPTM: VehicleID}
\label{fig:4b}
    \end{subfigure}\hfill
    \par\bigskip
    \begin{subfigure}{1.0\linewidth}
\includegraphics[trim=150 0 100 0,clip, width=1.0\linewidth]{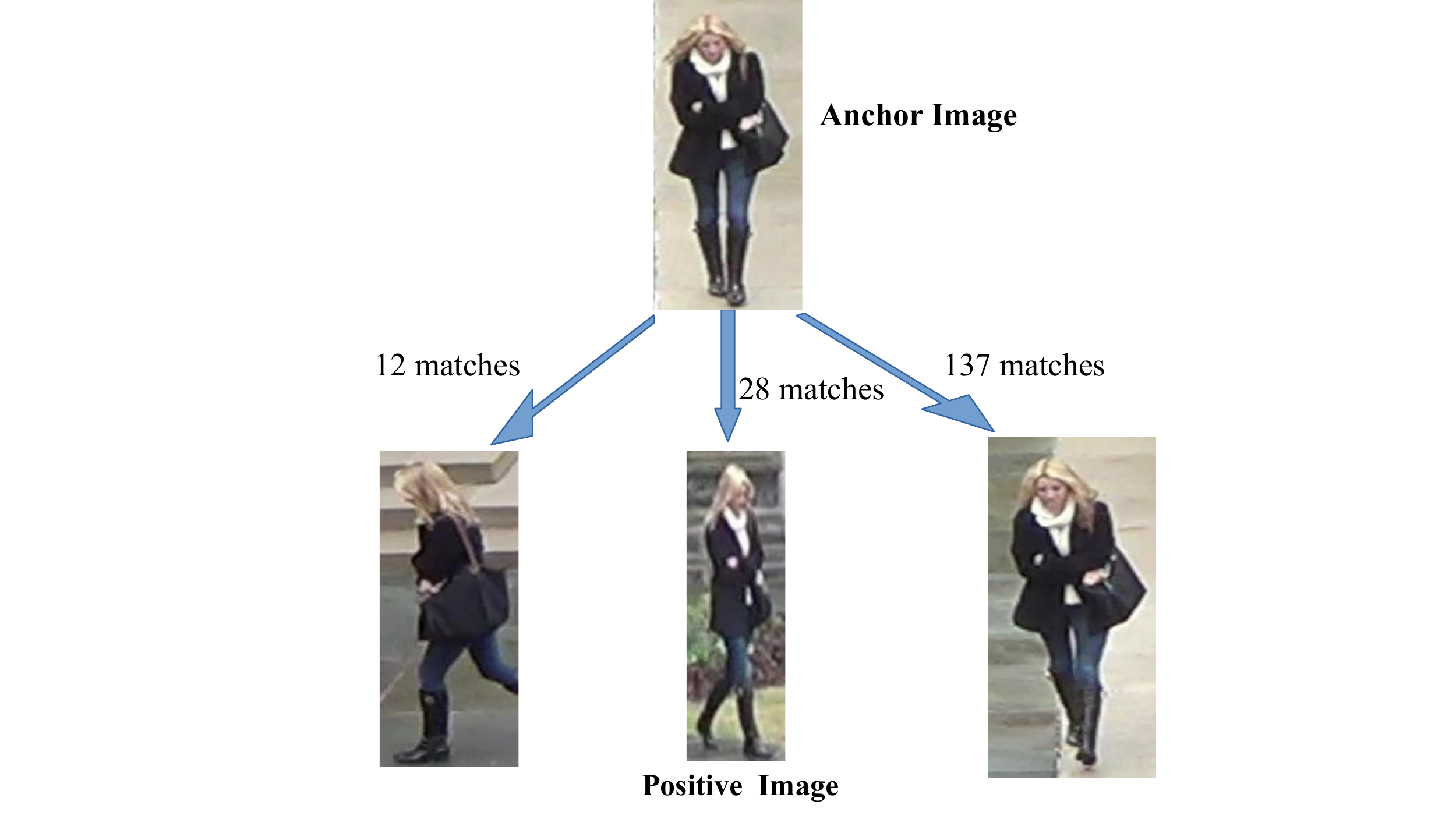} 
\caption{Anchor-Positive Pair Selection by RPTM: DukeMTMC}
\label{fig:4c}
    \end{subfigure}
\caption{
Visual representation of the Anchor-Positive Pair Selection methodology for Veri-776, VehicleID and DukeMTMC by the proposed RPTM algorithm, as explained in Section 4.2 of the main paper.
}
    \label{fig:pos}
\end{figure}

\section{Outlier Analysis}
\label{out}
Figure \ref{fig:prob} displays several cases of outliers in reID datasets. Outliers prevent images from mapping the correct semantic meaning of the IDs the images belong to, making outliers extremely problematic for reID. One of the most common outliers in reID datasets is images with occlusions that tend to mask the object being studied, wholly or partially. In person reID, bounding boxes sometimes capture two persons and the focus is put on the wrong object. In vehicle reID, especially for large datasets like VehicleID, a major issue is the presence of the same model across IDs. This outlier issue is further exacerbated when these models are of the same colour as well. All these cases make it difficult to train CNNs to learn accurate features. Hence, RPTM is proposed to exclude such outlier cases and learns additional features during positive selection for triplet mining. By cleaning up the reID process as described, RPTM can train reID models robust to outliers. 

\section{Triplet Mining and Triplet Loss}
\label{tr}
The triplet mining method proposed in Section 4.2 of the main paper is quite effective is selecting a suitable semi-hard positive sample for an anchor image during training. The triplet mining strategy used estimates threshold $\tau$ which reflects the bare minimum GMS matches $RPTM_{min}$, the non-zero average value of GMS matches $RPTM_{mean}$ or the maximum GMS matches $RPTM_{max}$, from which we select $RPTM_{mean}$ as the threshold for positive sample selection. Figure \ref{fig:pos} reflects the above strategy across all three benchmark datasets used in this paper.

Table \ref{tab:trip} describes the role of triplet loss. We have established how important positive mining is to the triplet loss cost function but it is prudent to evaluate the role triplet loss plays in reID. To that end, we manipulate the co-efficient of the triplet loss function $\lambda_{tri}$. Set at 2 for normal experiments, we change the value of $\lambda_{tri}$ to 0.5 and 1 and train the RPTM model on Veri-776 and DukeMTMC. Since cross-entropy loss is significantly larger than triplet loss, $\lambda_{xent}$ is set at 0.5 throughout our experiments.

\begin{table}[h]
\begin{center}
\begin{tabular}{p{3.4cm}|P{1cm}|P{1cm}|P{1cm}  }

& mAP &r = 1&r = 5\\
 \hline
Duke($\lambda_{tri}$=1) &80.6&86.8&94.1\\
Duke($\lambda_{tri}$=0.5) &77.2&83.8&92.6\\
\hline
Veri($\lambda_{tri}$=1)&81.8&94.7&96.9\\
Veri($\lambda_{tri}$=0.5)&79.9&93.4&95.8\\
 \hline
 \end{tabular}

\end{center}
\caption{Evaluation results with the $\lambda$ co-efficient for triplet-cost set at 0.5. We observe a significant drop in mAP and top-k rank across both datasets.}
 \label{tab:trip}
\end{table}

\section{Re-Ranking}
\label{rr}
Here, we test the variation of re-ranked mAP for the Veri-776 and DukeMTMC dataset. Implementing the process used by ~\cite{zhong2017re}, Figure \ref{fig:rr} involves manipulating and setting the values of three coefficients, $k_1, k_2$ and $\eta$, which represent co-efficients($k_1, k_2$) and penalty factors($\eta$) used to revise the ranking list calculated from standard evaluation and re-calculate the pairwise distance for the new ranking list. Figure \ref{fig:a} shows mAP results with the variation of $k1$ keeping $k2$ and $\eta$ at 15 and 0.2 for Veri-776 and at 10 and 0.2 for DukeMTMC. Figure \ref{fig:b} shows the impact of $k2$ on mAP, with $k1$ and $\eta$ set at 60 and 0.2 for Veri-776 and 20 and 0.2 for DukeMTMC. Finally, the impact of $\eta$ is studied in figure \ref{fig:c}, with $k1$ and $k2$ fixed at 60 and 15 for Veri-776 and at 20 and 10 for DukeMTMC respectively.

\begin{figure}[t]
\centering
    \begin{subfigure}{\linewidth}
\begin{tikzpicture}[trim left=-1.8em]
\begin{axis}[width=\textwidth,
    height=0.3\textheight,
    xlabel={k1},
    ylabel={mAP(\%)},
    xmin=0, xmax=100,
    ymin=82, ymax=90,
    xtick={0,20,40,60,80,100},
    ytick={82,84,86,88,90},
    legend pos=north east,
    ymajorgrids=true,
    grid style=dashed,
]

\addplot[
    color=red,
    mark=square,
    ]
    coordinates {
    (10,82.5)(15,83.8)(20,84.7)(25,85.2)(30,85.8)(35,86.2)(40,86.8)(50,87.2)(60,88)(70,87.6)(80,87.3)(90,86.9)(100,86.5)
    };
\addplot[
    color=green,
    mark=*,
    ]
    coordinates {
    (10,85.1)(15,87.5)(20,89.2)(25,88.8)(30,88.5)(35,88.1)(40,87.5)(50,86.9)(60,86.5)(70,86.2)(80,85.6)(90,85.1)(100,84.5)
    };
    \legend{Veri-776, DukeMTMC}
    
\end{axis}
\end{tikzpicture}
\caption{We fix $k2$ at 15(Veri)/10(Duke) and $\eta$ at 0.2. }
\label{fig:a}
    \end{subfigure}\hfill
    \begin{subfigure}{\linewidth}
\begin{tikzpicture}[trim left=-1.8em]
\begin{axis}[width=\textwidth,
    height=0.27\textheight,
    xlabel={k2},
    ylabel={mAP(\%)},
    xmin=0, xmax=50,
    ymin=80, ymax=90,
    xtick={0,10,20,30,40,50},
    ytick={80,82,84,86,88,90},
    legend pos=north east,
    ymajorgrids=true,
    grid style=dashed,
]

\addplot[
    color=red,
    mark=square,
    ]
    coordinates {
    (5,86.2)(10,87.3)(15,88)(20,87.4)(25,86.8)(30,85.8)(35,85.2)(40,84.7)(45,84.3)(50,83.8)
    };
\addplot[
    color=green,
    mark=*,
    ]
    coordinates {
    (5,87.6)(10,89.2)(15,88.1)(20,86.6)(25,86.1)(30,84.8)(35,83.8)(40,82.3)(45,81.5)(50,80.2)
    };
    \legend{Veri-776, DukeMTMC}

\end{axis}
\end{tikzpicture}
\caption{We fix $k1$ at 60(Veri)/20(Duke) and $\eta$ at 0.2.}
\label{fig:b}
    \end{subfigure}\hfill
    \begin{subfigure}{\linewidth}
\begin{tikzpicture}[trim left=-1.8em]
\begin{axis}[width=\textwidth,
    height=0.3\textheight,
    xlabel={$\eta$},
    ylabel={mAP(\%)},
    xmin=0, xmax=1,
    ymin=78, ymax=90,
    xtick={0,0.2,0.4,0.6,0.8,1.0},
    ytick={78,80,82,84,86,88,90},
    legend pos=north east,
    ymajorgrids=true,
    grid style=dashed,
]

\addplot[
    color=red,
    mark=square,
    ]
    coordinates {
    (0,87.4)(0.2,88)(0.4,86.5)(0.6,85.9)(0.8,83.4)(1.0,78)
    };
\addplot[
    color=green,
    mark=*,
    ]
    coordinates {
    (0,87.9)(0.2,89.2)(0.4,87.7)(0.6,86.9)(0.8,85.6)(1.0,79.5)
    };
    \legend{Veri-776, DukeMTMC}
    
\end{axis}
\end{tikzpicture} 
\caption{We fix $k1$ at 60(Veri)/20(Duke) and $k2$ at 15(Veri)/10(Duke).}
\label{fig:c}
    \end{subfigure}
\caption{Insight into the variation of mAP results with the manipulation of re-ranking parameters $k1,k2$ and $\eta$ for Veri-776(red) and DukeMTMC(green). Optimal values of parameters $k1$ and $k2$ vary with datasets, while the best mAP results are seen when $\eta$ is set to 0.2. }
\label{fig:rr}
\end{figure}

\section{More Experimental Results}
\label{exp}
\begin{figure*}[t]
\begin{center}
   \includegraphics[trim = 20 252 0 125,clip,width=1.0\linewidth]{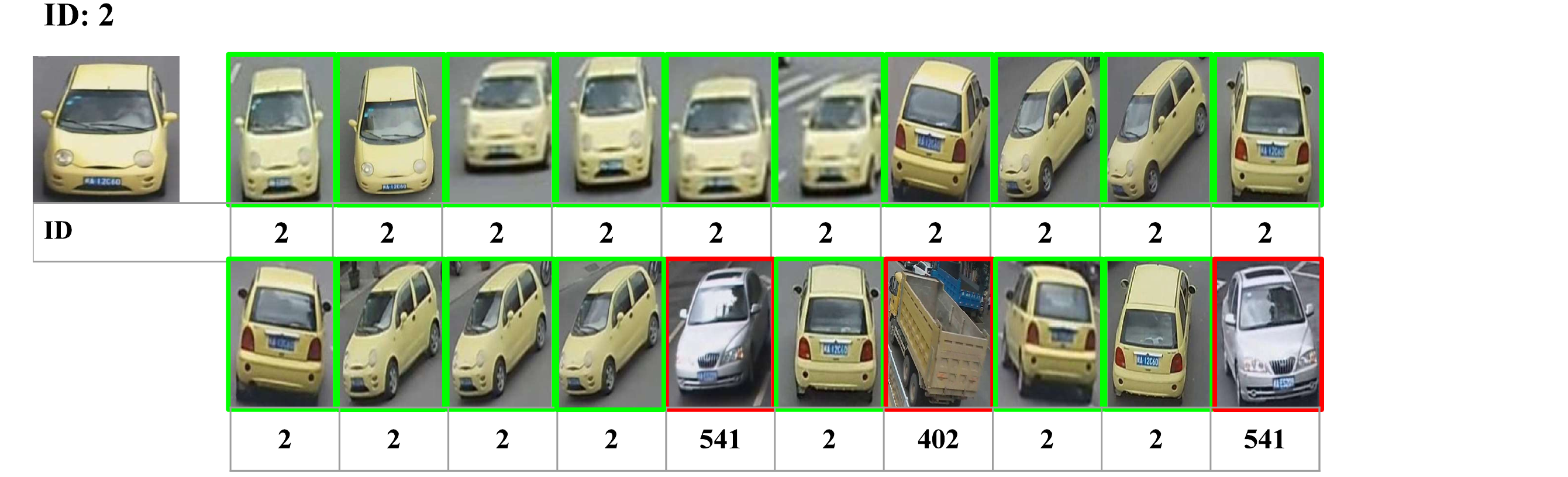}
   \put(-70, 70){DMT~\cite{he2020multi}}\\
   \includegraphics[trim = 20 252 0 128,clip,width=1.0\linewidth]{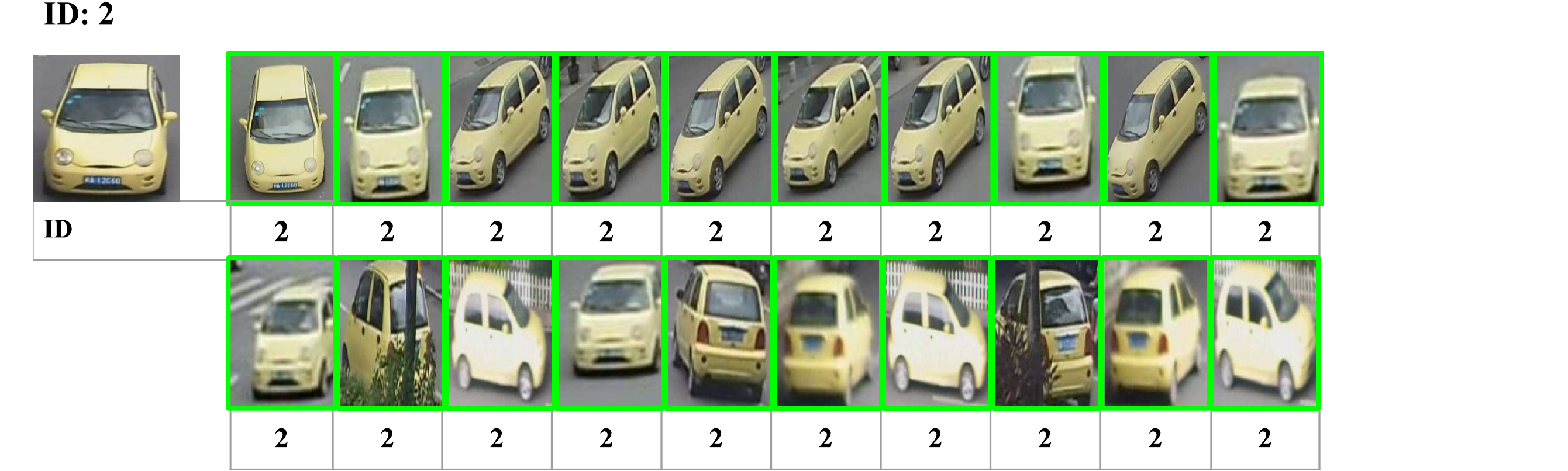}
   \put(-70, 70){RPTM(Ours)}\\
   \includegraphics[trim = 20 252 0 125,clip,width=1.0\linewidth]{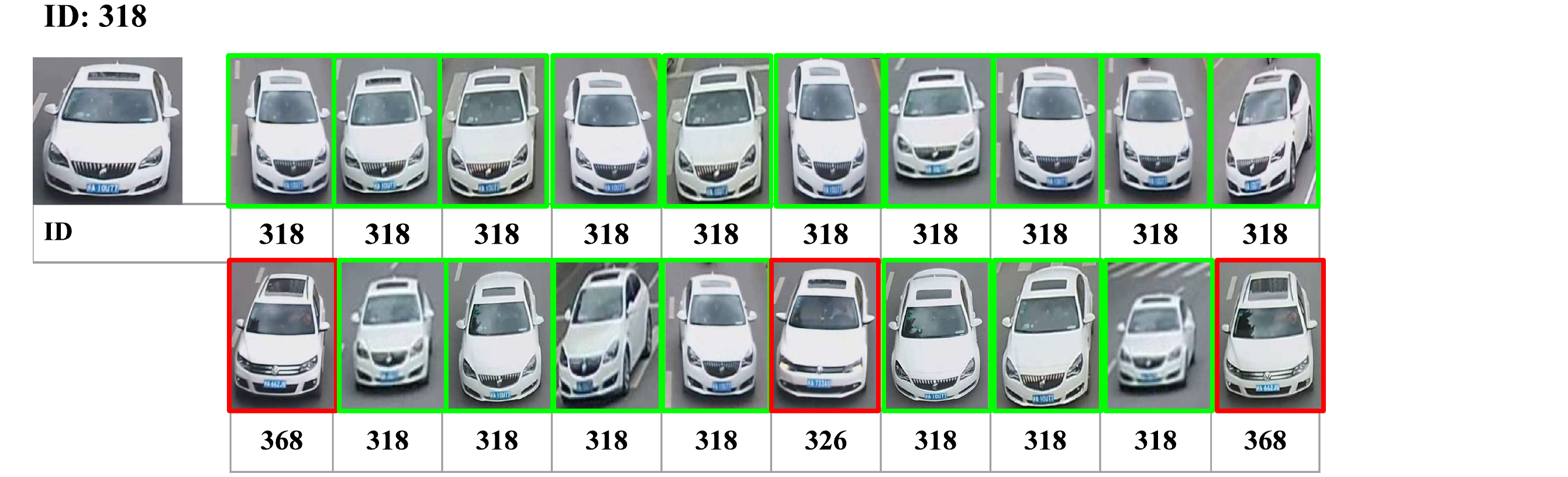}
   \put(-70, 70){DMT~\cite{he2020multi}}\\
   \includegraphics[trim = 20 252 0 128,clip,width=1.0\linewidth]{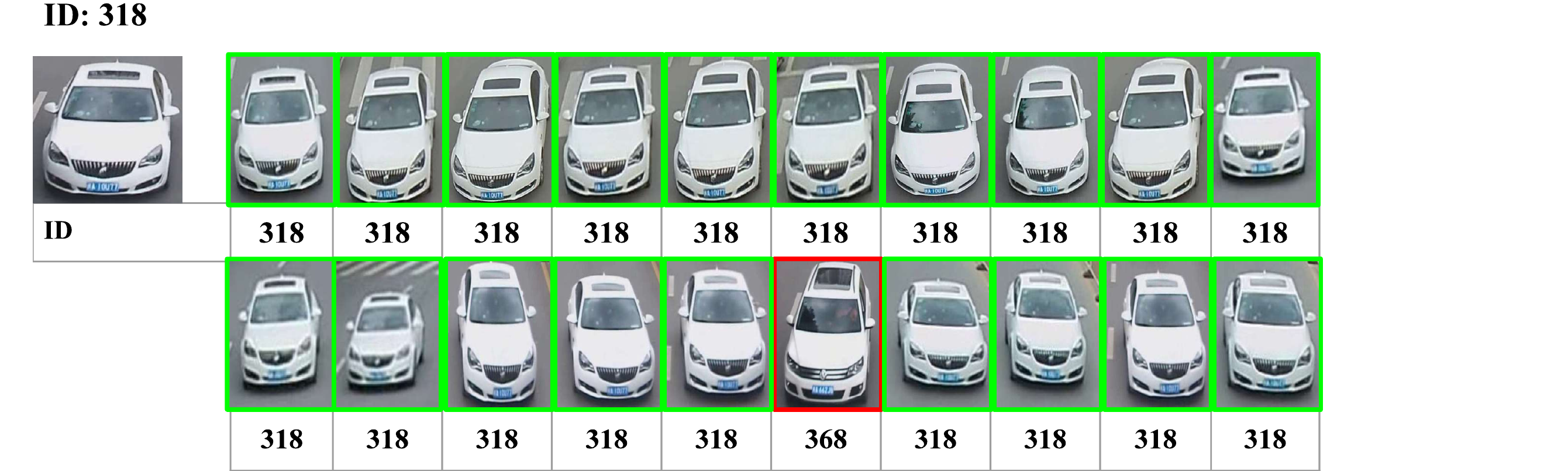}
   \put(-70,70){RPTM(Ours)}\\

\end{center}
   \caption{Example results of two query images from the Veri-776 dataset. We compare the top-20 gallery retrieval results between our proposed RPTM model and current state-of-the-art for Veri-776, DMT~\cite{he2020multi}. RPTM shows relatively better results throughout the retrieval task, but more notably in the top-11 to top-20 retrievals.}
\label{fig:veri}
\end{figure*}

\begin{figure*}[t]
\begin{center}
   \includegraphics[trim = 20 210 0 120,clip,width=1.0\linewidth, height =15.0em]{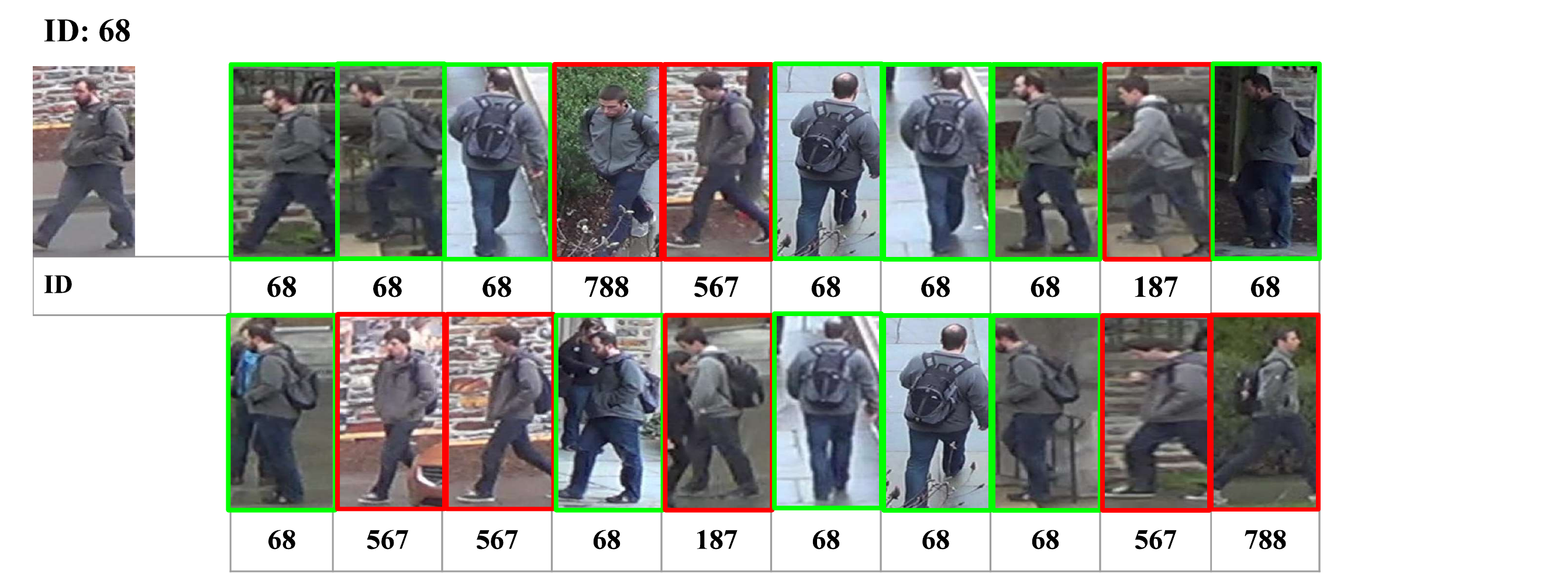}
   \put(-75, 70){st-ReID~\cite{wang2019spatial}}\\
   \includegraphics[trim =20 210 0 120,clip,width=1.0\linewidth, height =15.0em]{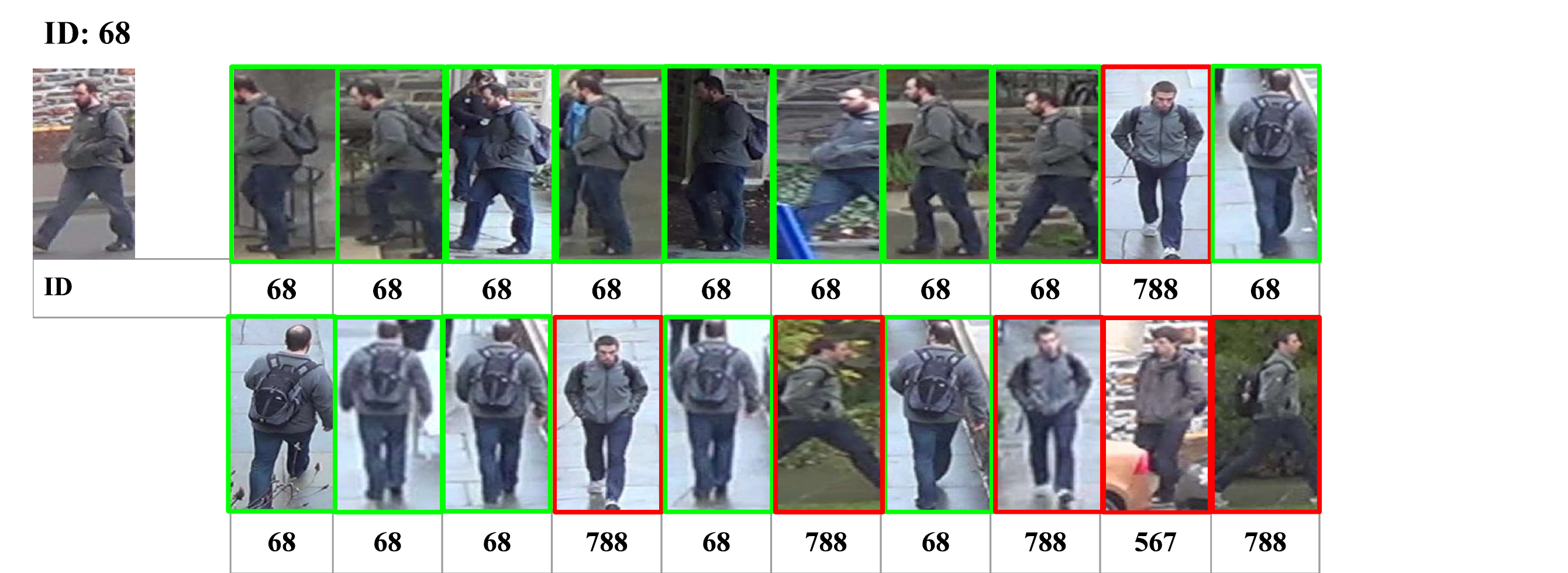}
   \put(-75, 70){RPTM(Ours)}\\
   \includegraphics[trim =20 210 0 120,clip,width=1.0\linewidth, height =15.0em]{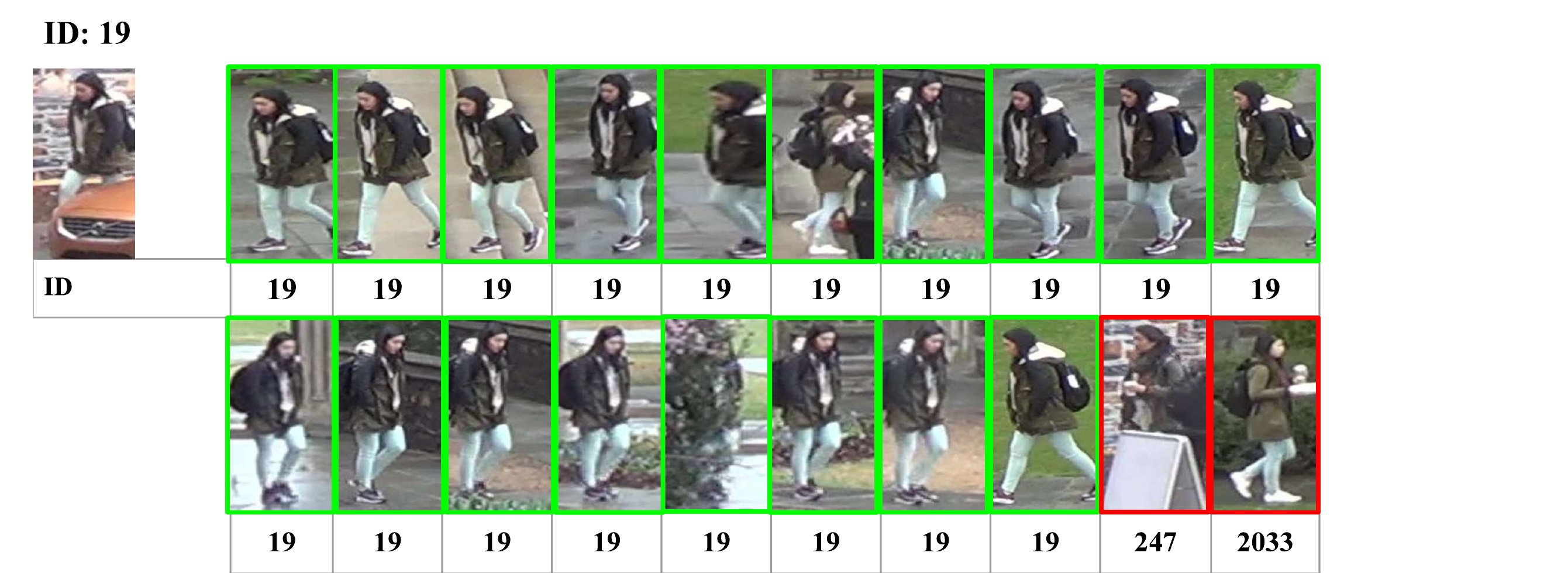}
   \put(-75, 70){st-ReID~\cite{wang2019spatial}}\\
   \includegraphics[trim =20 210 0 120,clip,width=1.0\linewidth, height =15.0em]{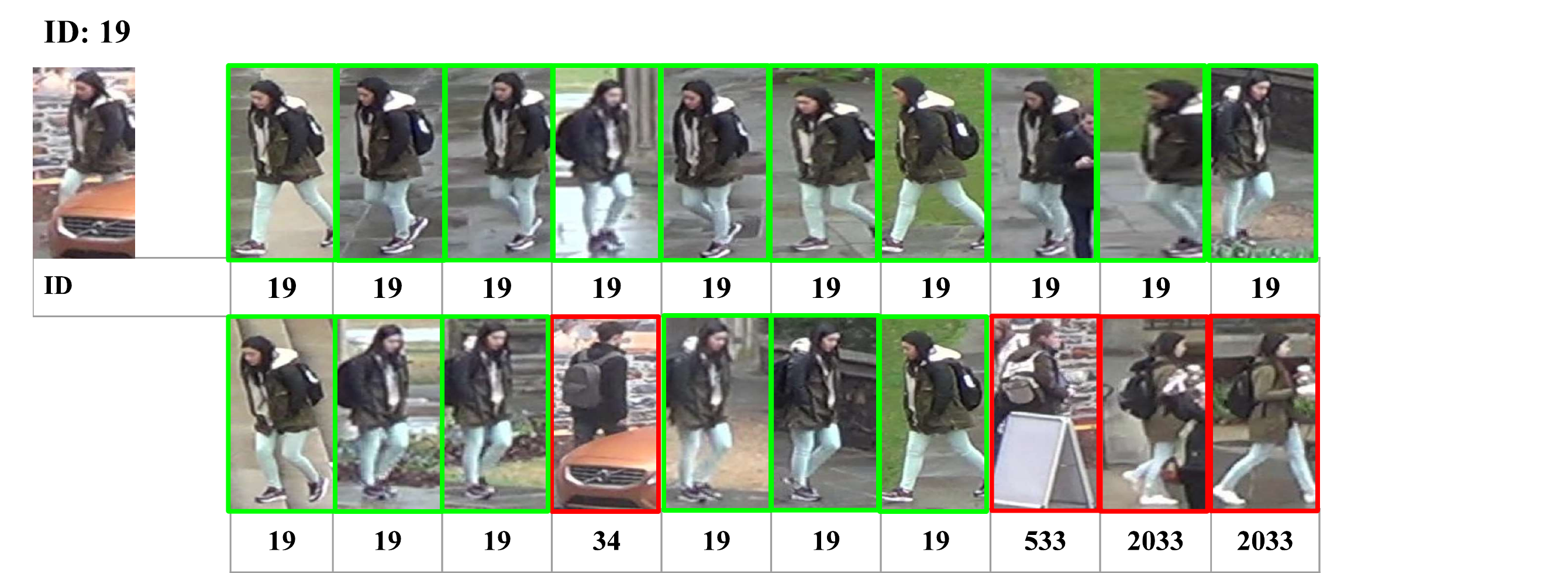}
   \put(-75,70){RPTM(Ours)}\\

\end{center}
   \caption{Top-20 retrieval results for two query images taken from DukeMTMC. We compare our RPTM model with the current state-of-the-art, st-ReID~\cite{wang2019spatial}. Despite st-ReID showing higher values of mAP, rank-1 and rank-5 results, and RPTM being fine-tuned for vehicle reID, RPTM shows equivalent retrieval results compared to st-ReID.}
\label{fig:duke}
\end{figure*}

In this section, we provide more visual comparisons with state-of-the-art methods, based on the top-k gallery matches for a given query sample, where we set the value of k to 20. Correct ID matches are denoted in green boxes whereas wrong matches are enclosed in red boxes. We compare our vehicle reID results on Veri-776 (Figure \ref{fig:veri}) with DMT~\cite{he2020multi} and our person reID results on DukeMTMC(Figure \ref{fig:duke}) with st-ReID~\cite{wang2019spatial}. The proposed algorithm generates strong pose-aware results, especially in the top-10 matches.

{\small
\bibliographystyle{ieee_fullname}
\bibliography{egbib}
}